%% file: iclr2025_conference.tex
\documentclass{article} 
\usepackage{iclr2025_conference,times}

\input{math_commands.tex}

\usepackage{hyperref}
\usepackage{url}
\usepackage{xcolor}
\usepackage{graphicx}
\usepackage{tcolorbox}
\usepackage{multirow}
\usepackage{pdfpages}
\usepackage[ruled,vlined]{algorithm2e}
\usepackage{algorithmic,amsmath}
\usepackage{wrapfig}
\usepackage{amssymb}
\usepackage{enumitem}
\usepackage{transparent}
\usepackage{booktabs}

\usepackage{subcaption}
\usepackage{ulem}
\usepackage{floatrow}
\newfloatcommand{capbtabbox}{table}[][\FBwidth]
\usepackage{blindtext}
\usepackage{listings}
\usepackage{tcolorbox} 
\tcbuselibrary{listings,skins} 
\definecolor{custompurple}{rgb}{0.408, 0.2, 0.6}  
\definecolor{customgreen}{rgb}{0.627, 0.808, 0.388}  
\definecolor{customblue}{rgb}{0.216, 0.329, 0.573}  
\title{ReDeEP: Detecting Hallucination in Retrieval-Augmented Generation via \mbox{Mechanistic Interpretability}}

\author{
  Zhongxiang Sun\textsuperscript{1}\thanks{Work done during their internship at Kuaishou.},  
  Xiaoxue Zang\textsuperscript{2},  
  Kai Zheng\textsuperscript{2},  
  Jun Xu\textsuperscript{1}\thanks{Corresponding author. Work partially done at Engineering Research Center of Next-Generation Intelligent Search and Recommendation, Ministry of Education.}\\
  \textbf{Xiao Zhang\textsuperscript{1}}, 
  \textbf{Weijie Yu \textsuperscript{3}},
  \textbf{Yang Song \textsuperscript{2}},
  \textbf{Han Li \textsuperscript{2}}
  \\
  \textsuperscript{1}Gaoling School of Artificial Intelligence, Renmin University of China, Beijing, China \\
  \textsuperscript{2}Kuaishou Technology Co., Ltd., Beijing, China \\
  \textsuperscript{3}School of Information Technology and Management, University of International Business and Economics
  \\
  \texttt{sunzhongxiang}@ruc.edu.cn \\
}

\iclrfinalcopy
\begin{document}

\maketitle

\begin{abstract}
Retrieval-Augmented Generation (RAG) models are designed to incorporate external knowledge, reducing hallucinations caused by insufficient parametric (internal) knowledge. However, even with accurate and relevant retrieved content, RAG models can still produce hallucinations by generating outputs that conflict with the retrieved information. Detecting such hallucinations requires disentangling how Large Language Models (LLMs) utilize external and parametric knowledge. Current detection methods often focus on one of these mechanisms or without decoupling their intertwined effects, making accurate detection difficult. In this paper, we investigate the internal mechanisms behind hallucinations in RAG scenarios. We discover hallucinations occur when the \textit{Knowledge FFNs} in LLMs overemphasize parametric knowledge in the residual stream, while \textit{Copying Heads} fail to effectively retain or integrate external knowledge from retrieved content. Based on these findings, we propose \textbf{ReDeEP}, a novel method that detects hallucinations by decoupling LLM’s utilization of external context and parametric knowledge. Our experiments show that ReDeEP significantly improves RAG hallucination detection accuracy. Additionally, we introduce AARF, which mitigates hallucinations by modulating the contributions of Knowledge FFNs and Copying Heads. 
\end{abstract}

\section{Introduction}
\label{sec:intro}
\begin{figure}[h]
    \centering
    \includegraphics[width=0.95\textwidth]{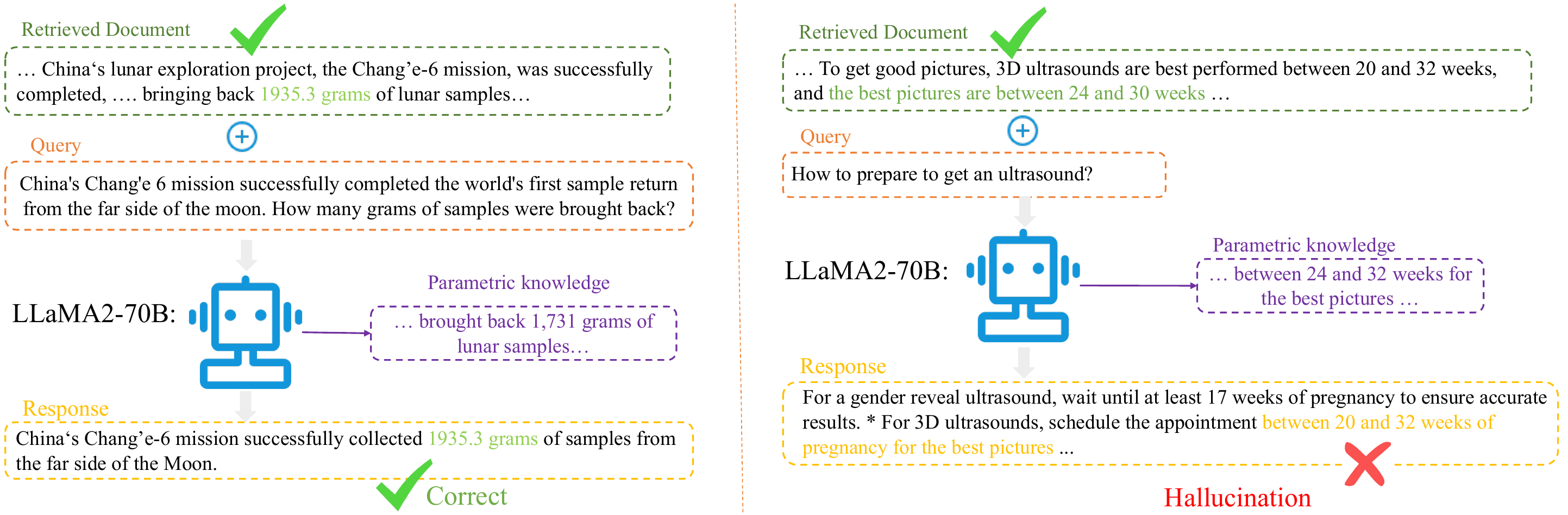}
    \caption{Two examples of RAG where the retrieved document is correct but conflicts with parametric knowledge. The left example shows a correct response based on external knowledge, while the right example demonstrates hallucination despite accurate external context.}
    \label{fig:introduction}
\end{figure}

LLMs have made significant advancements in natural language processing tasks~\citep{dubey2024llama,achiam2023gpt}. However, they still face challenges with hallucinations, often generating factually inaccurate outputs~\citep{huang2023survey}. To mitigate this issue, many researchers have introduced Retrieval-Augmented Generation (RAG) models, which aim to improve the accuracy of LLM responses by incorporating relevant information retrieved from external knowledge bases~\citep{shuster2021retrieval,gao2023retrieval}.

Despite the use of accurate and relevant retrieved context, RAG models may still produce statements that are either unsupported or contradict the retrieved information, a phenomenon we term \textbf{RAG Hallucination}~\citep{niu-etal-2024-ragtruth,magesh2024hallucination}. Recent studies have examined the potential conflicts between the \textbf{external context} and the LLM's \textbf{parametric knowledge} in RAG models~\citep{xu2024knowledge}. As shown in~\autoref{fig:introduction}, these conflicts can lead to hallucinations but do not always cause them. Therefore, it is important to distinguish RAG hallucination from Knowledge Conflict as a new research direction. \emph{Our work focuses on detecting RAG hallucinations, specifically in cases where the retrieved external context is accurate and relevant.}

\begin{figure}[t]
    \centering
    \includegraphics[width=0.98\textwidth]{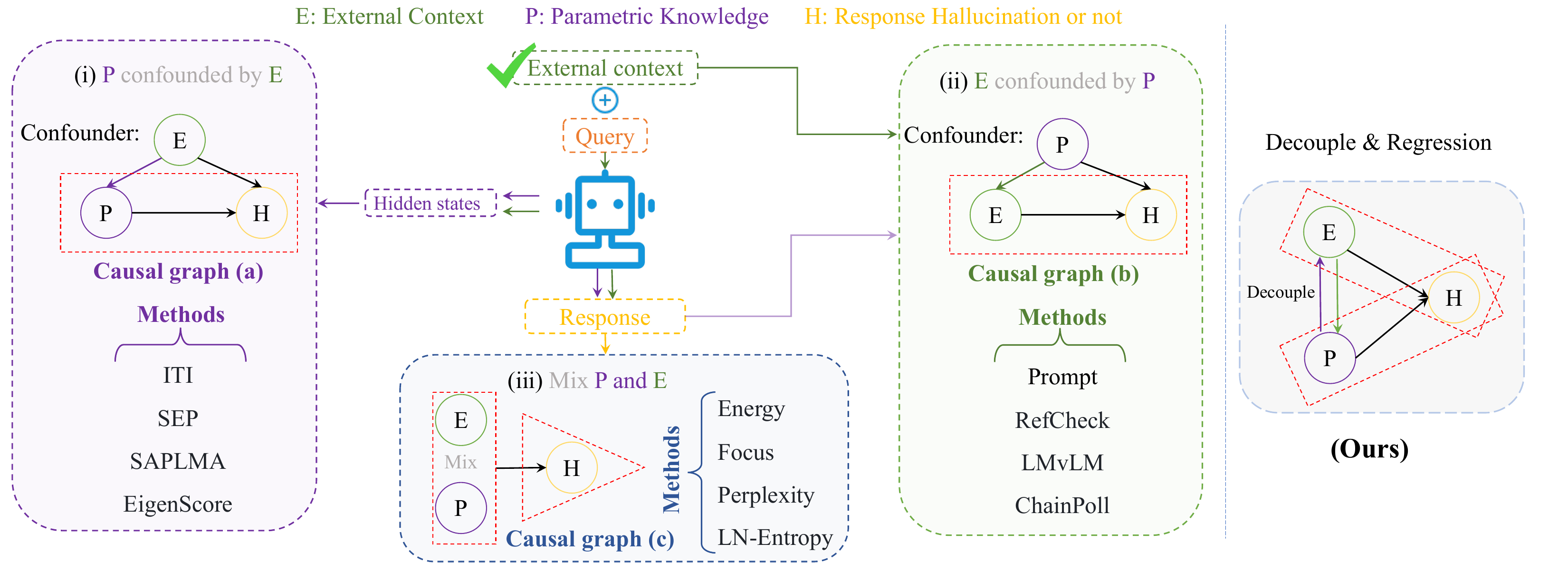}
    \caption{Causal perspectives on hallucination detection methods. \textcolor{custompurple}{(i)}: parametric knowledge is confounded by external context, \textcolor{customgreen}{(ii)}: external context is confounded by parametric knowledge, and \textcolor{customblue}{(iii)}: mixes both without decoupling their contributions. \textbf{(Ours)}: decouple these confounders using mechanistic interpretability, incorporating them as covariates to improve hallucination detection.}
    \label{fig:related_work}
\end{figure}

Existing hallucination detection methods can be categorized into three causal frameworks~\citep{neuberg2003causality,pearl2009causality}, as illustrated in~\Figref{fig:related_work} (detailed introduction of methods see Appendix~\ref{sec:baseline_de}): (i) \textbf{Parametric Confounded by External:} which relies on the LLM's hidden states for hallucination detection, where the external context ($\ermE$) serves as a confounder between parametric knowledge ($\ermP$) and hallucinations ($\ermH$). From a knowledge storage perspective~\citep{geva2021transformer}, hidden states represent the result of querying the parametric knowledge ($\ermP$) with external context ($\ermE$), establishing a causal path from $\ermE$ to $\ermP$ (\textcolor{custompurple}{graph (a)}). The presence of $\ermE$ as a confounder complicates the accurate prediction of hallucinations based on $\ermP$ alone~\citep{chyzhyk2022remove}. (ii) \textbf{External Confounded by Parametric:} which focuses on hallucination detection by leveraging external context and model responses. Here, parametric knowledge ($\ermP$) is a confounder between the external context ($\ermE$) and hallucinations ($\ermH$), creating a causal link from $\ermP$ to $\ermE$ (\textcolor{customgreen}{graph (b)}) due to the unavoidable presence of parametric knowledge in the response. (iii) \textbf{Mixed Parametric and External:} which combines both parametric and external knowledge directly, often using uncertainty or sampling techniques (e.g., token probability) to detect hallucinations~(\textcolor{customblue}{graph (c)}). However, this mixing of $\ermE$ and $\ermP$ without decoupling their roles obscures their individual contributions~\citep{bengio2013representation}.

To address the challenges of hallucination detection in RAG models, we first leverage mechanistic interpretability~\citep{ferrando2024primer,transformercircuits2021} to decouple the LLM's utilization of parametric knowledge and external context. Specifically, we conduct an empirical study to explore the internal mechanisms behind hallucination generation in RAG scenarios. We introduce two metrics: the \textbf{External Context Score}, which uses attention heads to quantify the model’s utilization on external context, and the \textbf{Parametric Knowledge Score}, which is based on FFNs to evaluate LLM’s utilization of parametric knowledge (\S~\ref{sec:metrics}). Correlation analysis and causal intervention reveals that hallucinations typically occur when \textbf{Knowledge FFNs} (from later LLM layers) over-add parametric knowledge into the residual stream, while \textbf{Copying Heads} (attention heads exhibiting copying behaviours) neglect the necessary external knowledge from retrieved content or LLM loses the information attended to by Copying Heads during the generation process (\S~\ref{sec:rq1}).

Building on our causal analysis and mechanistic interpretability, we propose \textbf{ReDeEP} (\textbf{Re}gressing \textbf{De}coupled \textbf{E}xternal context score and \textbf{P}arametric knowledge score) for detecting hallucinations in LLM-based RAG, which treat parametric knowledge ($\ermP$) and external context ($\ermE$) as covariates to solve the confounding problem~\citep{kahlert2017control} (see \textbf{Ours} in~\Figref{fig:related_work}).  Additionally, we introduce \textbf{AARF} (\textbf{A}dd \textbf{A}ttention \textbf{R}educe \textbf{F}FN), which mitigates hallucinations by modulating the contributions of Knowledge FFNs and Copying Heads in the residual stream (\S~\ref{sec:methods}). Experiments on RAGTruth and Dolly (AC) confirm that ReDeEP significantly outperforms existing detection methods, while AARF improves the truthfulness of LLaMA models (\S~\ref{sec:exp}).


\section{Background and Related Works}
\subsection{Background}
\label{sec:back}
Our work is grounded in mechanistic interpretability~\citep{ferrando2024primer,nostalgebraist2020,meng2022locating,transformercircuits2021}, which aims to explain how individual components of language models (LMs) contribute to predictions. In this study, we focus on transformer decoder-only architectures (GPT-like models) due to their widespread use~\citep{achiam2023gpt,dubey2024llama}. Transformers use residual connections, where each layer adds information from \textit{Attention Heads} and \textit{Feed-Forward Networks} (FFNs) to the hidden state via the residual stream, contributing to the final prediction~\citep{transformercircuits2021}.

\textbf{Attention Heads:} Attention heads play a crucial role in contextualizing token representations by selectively attending to previous tokens and updating the residual stream~\citep{ferrando2024information,clark-etal-2019-bert,wu2024retrieval}. Notably, some attention heads, referred to as \textit{Copying Heads}, have been shown to copy information from one token to another through their OV (output-value) circuits~\citep{transformercircuits2021}. These heads can be identified by analyzing the positive eigenvalues of the OV matrix, which indicate copying behavior. Copying Heads contributes to preserving previously attended tokens in the residual stream, which is critical for external context utilization.

\textbf{FFNs:} FFN layers primarily function as knowledge storage in transformers~\citep{geva2021transformer}. Each FFN layer transforms the hidden state by linearly combining key-value pairs, where keys encode specific knowledge and values represent the output of this knowledge.  Research shows that FFNs are critical for the utilization of parametric knowledge within LLMs, enabling the model to retrieve and integrate stored information effectively for prediction~\citep{dai2022knowledge}.

\textbf{Logit Lens:} The \(\operatorname{LogitLens}\) is a technique that decodes hidden states \(\boldsymbol{x}^{l}\) directly into the vocabulary distribution using the LayerNorm and the unembedding matrix $\boldsymbol{W}_U$ of the LLM for interpretability~\citep{nostalgebraist2020}:
\begin{equation}
    \operatorname{LogitLens}\left(\boldsymbol{x}^{l}\right) = \operatorname{LayerNorm}(\boldsymbol{x}^{l}) \boldsymbol{W}_U
    \label{eq:logit_lens}.
\end{equation}

By understanding the roles of Attention Heads (e.g., Copying Heads) and FFNs, we can better interpret the internal states of LLMs and identify the mechanisms behind hallucinations in RAG scenarios. Detailed background information can be found in Appendix~\ref{sec:back_details}.

\subsection{Related Work}


\textbf{Hallucination of LLMs:} LLMs often generate hallucinations—content inconsistent with real-world facts or inputs~\citep{huang2023survey}. As depicted in Figure~\ref{fig:related_work}, although there has been extensive research on detecting hallucinations~\citep{niu-etal-2024-ragtruth,manakul2023selfcheckgpt,han2024semantic}, few studies have concentrated on RAG hallucinations, particularly on the internal mechanisms driving these hallucinations. Inspired by research on Knowledge Conflicts~\citep{xu2024knowledge}, our work is the first to apply mechanistic interpretability to lens the internal mechanisms of RAG hallucinations from the perspectives of LLM's utilization of external and parametric knowledge, leading to a more accurate detection method than previous approaches.

\textbf{Mechanistic Interpretability:} Mechanistic interpretability~\citep{ferrando2024primer,transformercircuits2021} seeks to explain the internal processes of LLMs, enabling the interpretation of how individual model components contribute to the final prediction. Our work builds on insights into FFN layers, attention heads~\citep{vaswani2017attention}, residual streams~\citep{transformercircuits2021}, and the logit lens~\citep{nostalgebraist2020} to analyze the internal mechanisms of LLMs when RAG hallucinations occur.

\section{Empirical study}
\label{sec:empirical}





Our empirical study investigates how hallucinations in RAG models relate to the internal states of the LLM. Using Mechanistic Interpretability techniques, we focus on how the LLM’s use of external context and parametric knowledge contributes to hallucinations.

\begin{figure}[t]
    \centering
    \includegraphics[width=\textwidth]{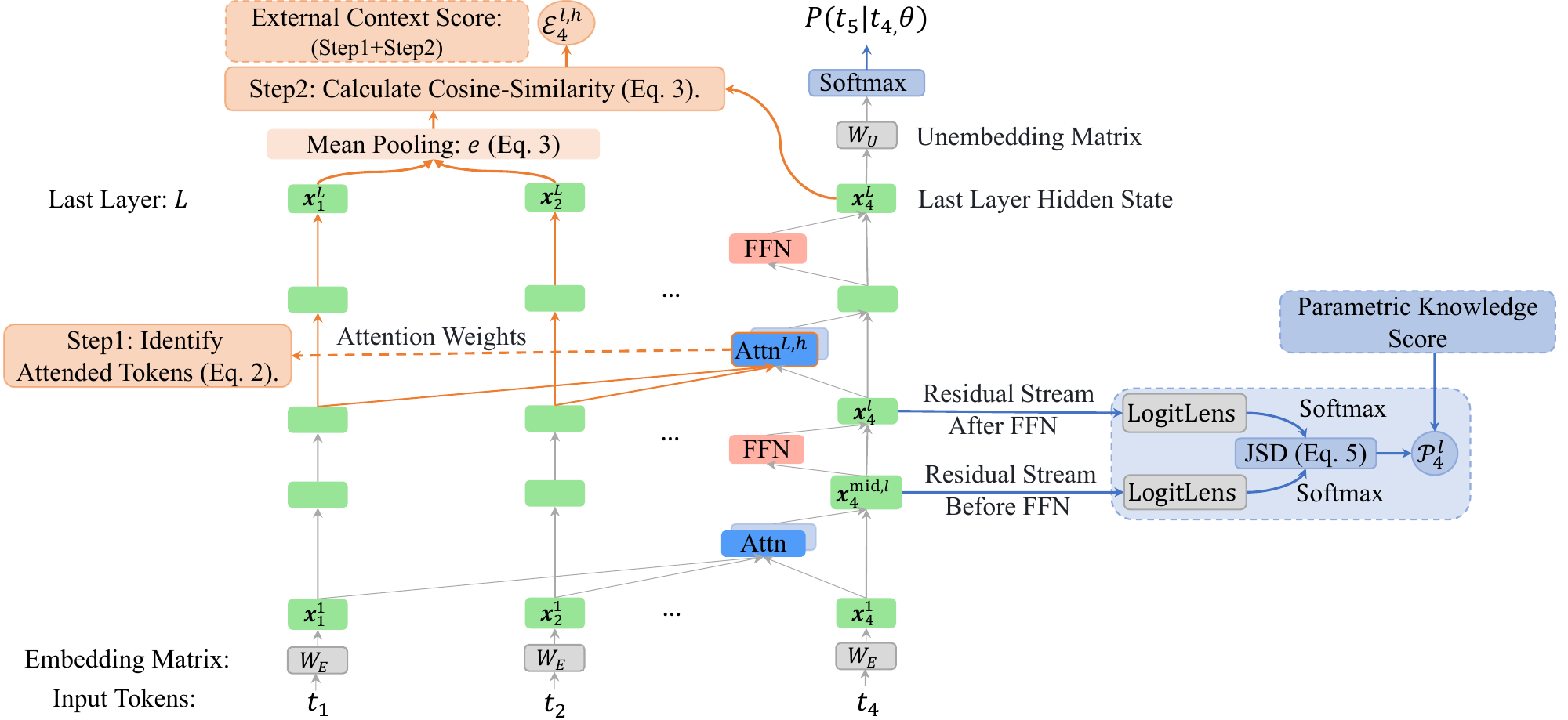}
    \caption{Expanded views of Unrolled LLMs' Attention and FFN blocks. (a): The calculation process of the External Context Score and Parametric Knowledge Score. (b): Example of intervening on attention heads. (c): Example of intervening on FFN modules.} 
    \label{fig:method}
\end{figure}

\textbf{Experiment Setting:}
We conduct experiments on the Llama2-7B-chat model~\citep{touvron2023llama} using the training set of RAGTruth dataset~\citep{niu-etal-2024-ragtruth}, a high-quality, manually annotated dataset for RAG hallucinations (Details in~\Secref{sec:setting}). Each data point in RAGTruth consists of a query \(\mathbf{q}\), retrieved context \(\mathbf{c}\), response \(\mathbf{r}\), and the hallucination label \(h\) (where 0 is truth and 1 is hallucination). 
During generation, the input to the LLM \(f\) is a sequence of tokens \(\mathbf{t} = \left\langle t_1, t_2, \ldots, t_n \right\rangle\), including the query \(\mathbf{q} = \left\langle t_{1}, \ldots, t_{q} \right\rangle\), retrieved context \(\mathbf{c} = \left\langle t_{q+1}, \ldots, t_{c} \right\rangle\), and a partial generated response \(\hat{\mathbf{r}} = \left\langle t_{c+1}, \ldots, t_{n} \right\rangle\).

\subsection{Metrics for LLMs' Utilization of External Context and Parametric Knowledge}
\label{sec:metrics}
To quantify how LLMs use external context and parametric knowledge, we design two specific metrics, as shown in \Figref{fig:method} (a):

\textbf{External Context:}  Considering attention heads primarily function to retrieve relevant information (as discussed in \Secref{sec:back}), we measure the LLM’s utilization of external context by assessing (1) whether attention heads focus on the correct context, and (2) whether the LLM effectively retains and utilizes this information during generation. To evaluate these aspects, we define the following metric based on the semantic difference between the external context attended by attention heads and the generated information:

For the last token \(t_n\), the attention weights on the context are \(\boldsymbol{a}^{l, h}_{n, q:c}\), where \(\boldsymbol{a}_{n}^{l,h}\) is obtained from~\Eqref{eq:attention_weight}. We select the top \(k\%\) tokens with the highest attention scores as attended tokens:
\begin{equation}
    \mathcal{I}_{n}^{l,h} = \operatorname{arg\,top}_{k\%}(\boldsymbol{a}_{n,q:c}^{l,h}).
\end{equation}
Given that attention often shows high sparsity~\citep{zhu2024near,zhang2024h2o}, with only a few tokens capturing most of the attention scores, we choose \( k = 10 \) to ensure coverage of high-attention tokens and maintain token diversity.

Inspired by~\citep{luo2024foundations, chen2024inside}, which validated that the hidden states of LLM can serve as token semantic representations, we compute the token-level \textbf{E}xternal \textbf{C}ontext \textbf{S}core (ECS) based on the cosine-similarity between the mean-pooling of the last layer hidden states of attended tokens and the hidden state of token \(t_n\):
\begin{equation}
    \mathcal{E}_{n}^{l,h} = \frac{\boldsymbol{e} \cdot \boldsymbol{x}^{L}_{n}}{\|\boldsymbol{e}\| \|\boldsymbol{x}^{L}_{n}\|}, \quad \boldsymbol{e} = \frac{1}{|\mathcal{I}_{n}^{l,h}|} \sum_{j \in \mathcal{I}_{n}^{l,h}} \boldsymbol{x}^{L}_{j}.
\end{equation}
The response-level ECS is the average of token-level scores:
\begin{equation}
    \mathcal{E}^{l,h}_{\mathbf{r}} =  \frac{1}{|\mathbf{r}|}\sum_{t \in \mathbf{r}} \mathcal{E}^{l,h}_{t}.
    \label{eq:external_score}
\end{equation}

\textbf{Parametric Knowledge:} Considering FFNs store parametric knowledge, to assess how LLM use Parametric Knowledge (as discussed in \Secref{sec:back}), we use the \(\operatorname{LogitLens}\) to map residual stream states before (i.e., \(\boldsymbol{x}_{n}^{\mathrm{mid},l}\), calculated from~\Eqref{eq:x_mid}) and after the FFN layer (i.e., \(\boldsymbol{x}_{n}^{l}\), calculated from~\Eqref{eq:x_final}) to vocabulary distributions. The difference in vocabulary distributions represents the parametric knowledge added by the FFN layer to the residual stream, which is measured by Jensen-Shannon divergence (JSD), gives the token-level \textbf{P}arametric \textbf{K}nowledge \textbf{S}core (PKS):
\begin{equation}
    \mathcal{P}_{n}^{l} = \operatorname{JSD}\left(q(\boldsymbol{x}_{n}^{\mathrm{mid},l}) \parallel q(\boldsymbol{x}_{n}^{l})\right),
\end{equation}
where \(q(\boldsymbol{x}) = \operatorname{softmax}(\operatorname{LogitLens}(\boldsymbol{x}))\). 
The response-level PKS is the average of token-level scores:
\begin{equation}
    \mathcal{P}^{l}_{\mathbf{r}} =  \frac{1}{|\mathbf{r}|}\sum_{t \in \mathbf{r}} \mathcal{P}^{l}_{t}.
    \label{eq:parametric_score}
\end{equation}


Although these metrics may not be exact due to the complexity of LLMs, they serve as intuitive proxies that are sufficiently aligned with the understanding of existing works to analyze the LLM's use of external context and parametric knowledge in relation to hallucinations, as explored in the following questions.

\begin{figure}[t]
    \centering
    \includegraphics[width=\textwidth]{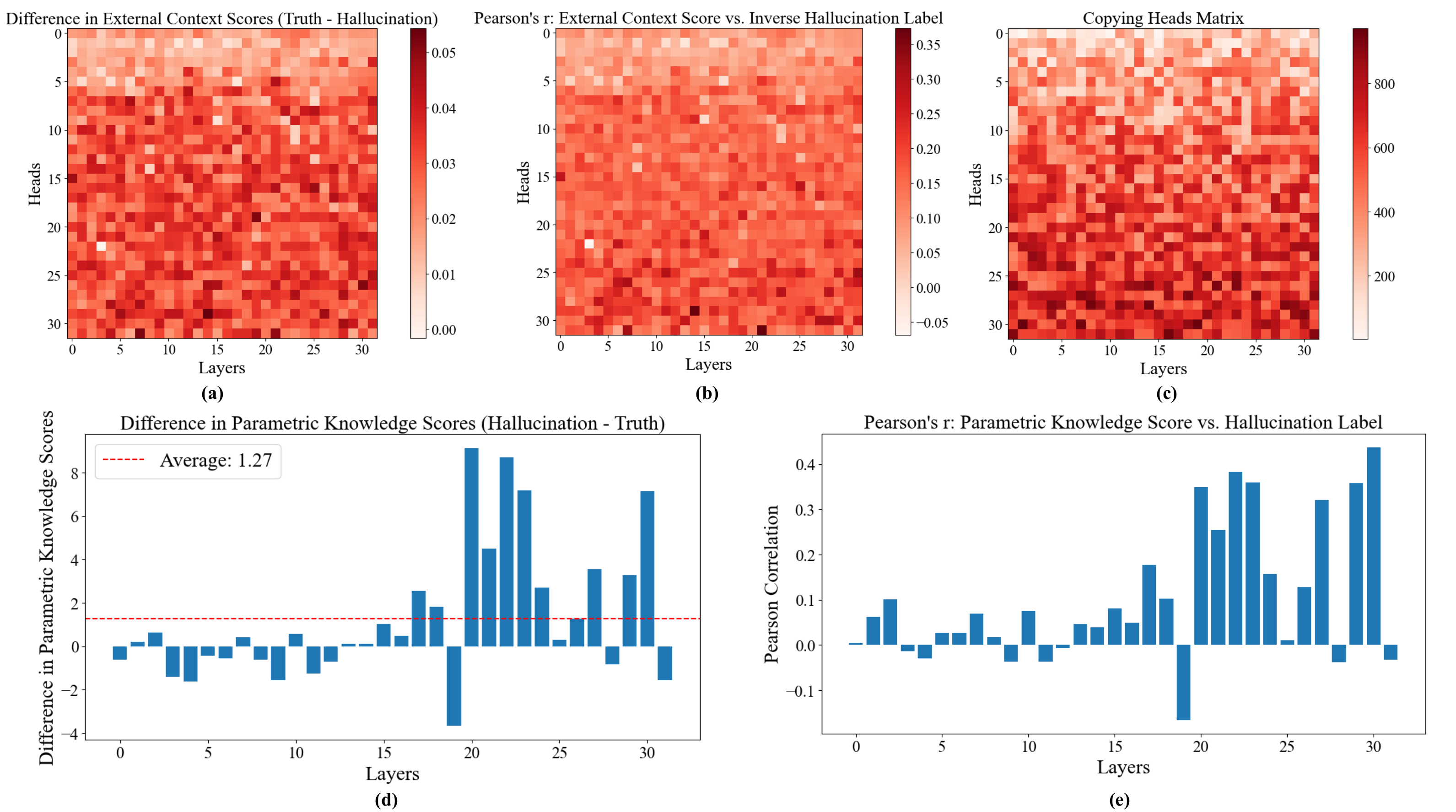}
    \caption{\textbf{Relationship Between LLM Utilization of External Context, Parametric Knowledge, and Hallucinations.} \emph{Top} shows the internal mechanism of LLM's utilization of external context and the occurrence of hallucinations, where the Pearson correlation coefficient between (c) and (a) is 0.41, and between (c) and (b) is 0.46, indicating correlations among them. \emph{Bottom} illustrates the internal mechanism of LLM's utilization of parametric knowledge and the occurrence of hallucinations, where (d) is scaled by $1e^{7}$.}
    \label{fig:empirical_1}
\end{figure}
\subsection{Experiments}
\label{sec:rq1}

\emph{\textbf{RQ1:} Relationship Between LLM Utilization of External Context, Parametric Knowledge, and Hallucinations}

\hypertarget{rq1-1}{\textit{(1) We first analyze the relationship between \textbf{External Context and RAG Hallucinations}:}}

\textbf{ECS Differences between Truthful and Hallucinated Responses:}  To investigate the relationship between LLM utilization of external context and hallucinations, we compare the external context score \(\mathcal{E}\) between truthful responses (\(h = 0\)) and hallucinated responses (\(h = 1\)). Specifically, we construct two subsets from the dataset: \(\mathcal{D}^{H}\) for hallucinations (\(h = 1\)) and \(\mathcal{D}^{T}\) for truthful responses (\(h = 0\)), and calculate the external context score difference for different attention heads:

\[
\Delta \mathcal{E}^{l,h} = \mathcal{E}_{T}^{l,h} - \mathcal{E}_{H}^{l,h} = \frac{1}{|\mathcal{D}^{T}|}\sum_{\mathbf{r} \in \mathcal{D}^{T}} \mathcal{E}^{l,h}_{\mathbf{r}} - \frac{1}{|\mathcal{D}^{H}|}\sum_{\mathbf{r} \in \mathcal{D}^{H}} \mathcal{E}^{l,h}_{\mathbf{r}}.
\]

\textbf{Result:} As shown in~\Figref{fig:empirical_1}(a), in Llama2-7B, 1006 out of 1024 attention heads show higher external context scores on the truthful dataset \(\mathcal{D}^{T}\) compared to the hallucination dataset \(\mathcal{D}^{H}\) (i.e., \(\Delta \mathcal{E}^{l,h} > 0\)). Since the external context score represents the LLM's utilization of external context through attention heads, we can conclude that, at a group level, LLMs utilize external context information less than truthful responses when generating hallucinations.

\textbf{Correlation between ECS and Hallucination:} To examine whether neglecting external context relates to RAG hallucinations, we analyzed the Pearson Correlation Coefficient (PCC) between the hallucination label and the external context score across data points in \(\mathcal{D}\). Given the expected negative correlation, we inverted the hallucination label \(h\) (denoted as \(\bar{h}\)) and used PCC to quantify the relationship between \(\{\bar{h}_{i}\}_{i=1}^{N}\) and external context scores \(\{\mathcal{E}_{i}\}_{i=1}^{N}\).

\textbf{Result:} As shown in~\Figref{fig:empirical_1}(b), most attention heads show negative correlation between external context scores and hallucination labels \(h\). Since the external context score indicates LLMs' utilization of external context, \Figref{fig:empirical_1}(a) and (b) suggest that RAG hallucinations occur when the LLM inadequately leverages external context. Further analysis (Appendix~\ref{sec:dive_ecs}) shows that hallucinations stem primarily from the LLM losing information attended by attention heads during generation rather than attention heads neglecting external knowledge.



\textbf{Relation between Copying Heads and Hallucination:} We observed that the external context score \(\mathcal{E}^{l,h}\) of certain attention heads correlates strongly with hallucinations, prompting further exploration of these heads' characteristics. Inspired by the Copying Heads concept from \Secref{sec:back}, we examined the relationship between these heads and Copying Heads. The calculation process of each attention head’s copying head score \(\mathcal{C}^{l,h}\) is shown in Appendix~\ref{sec:cal_copy_heads}).

\textbf{Result:} As shown in~\Figref{fig:empirical_1}(c), the correlation with the results in~\Figref{fig:empirical_1}(a) and (b) indicates that attention heads associated with hallucinations are often Copying Heads (PCC between (c) and (a) is 0.41, and (c) and (b) is 0.46). When these Copying Heads have low external context scores, they either fail to attend to the correct external context or, if attended, fail to retain and utilize this information effectively. This reduces the LLM's copying ability and leads to hallucinations, explaining the negative correlation between these heads' external context scores and the hallucination label \(h\).

\hypertarget{rq1-2}{{\textit{(2) Next, we analyze the relationship between \textbf{Parametric Knowledge and RAG Hallucinations}:}}}

\textbf{PKS Differences between Truth and Hallucination:} We compare the Parametric Knowledge Score \(\mathcal{P}\) across different layers when the LLM generates hallucinations versus truthful responses:

\[
\Delta \mathcal{P}^{l} = \mathcal{P}_{H}^{l} - \mathcal{P}_{T}^{l} = \frac{1}{|\mathcal{D}^{H}|}\sum_{\mathbf{r} \in \mathcal{D}^{H}} \mathcal{P}^{l}_{\mathbf{r}} - \frac{1}{|\mathcal{D}^{T}|}\sum_{\mathbf{r} \in \mathcal{D}^{T}} \mathcal{P}^{l}_{\mathbf{r}}.
\]

\textbf{Result:} As shown in~\Figref{fig:empirical_1}(d), parametric knowledge scores in the later layers of FFN modules are significantly higher in the hallucination dataset compared to the truthful dataset (i.e., \(\Delta \mathcal{P}^{l} > 0\)). On average, across all layers, hallucination responses exhibit higher parametric knowledge scores than truthful ones.

\textbf{Correlation between PKS and Hallucination:} To further explore the relationship between parametric knowledge and hallucinations, we calculate the Pearson correlation between the hallucination label and parametric knowledge scores.

\textbf{Result:} As shown in~\Figref{fig:empirical_1}(e), parametric knowledge scores in the later layers' FFN modules are positively correlated with the hallucination label \(h\) and we define the FFN modules from later layers that show strong correlations with hallucinations as \textbf{Knowledge FFNs}.  Since these scores represent the amount of parametric knowledge added to the residual stream, we conclude excessive addition of parametric knowledge by these Knowledge FFNs leads to hallucinations. This aligns with findings from LLM early exit studies~\citep{chuang2024dola,schuster2022confident}: when external context provides sufficient information, shallow layers can generate truthful responses, but over-reliance on parametric knowledge from deeper layers can confuse the model, causing hallucinations.

\begin{figure}[t]
    \centering
    \begin{minipage}[t]{0.48\textwidth}
        \centering
        \includegraphics[width=\textwidth]{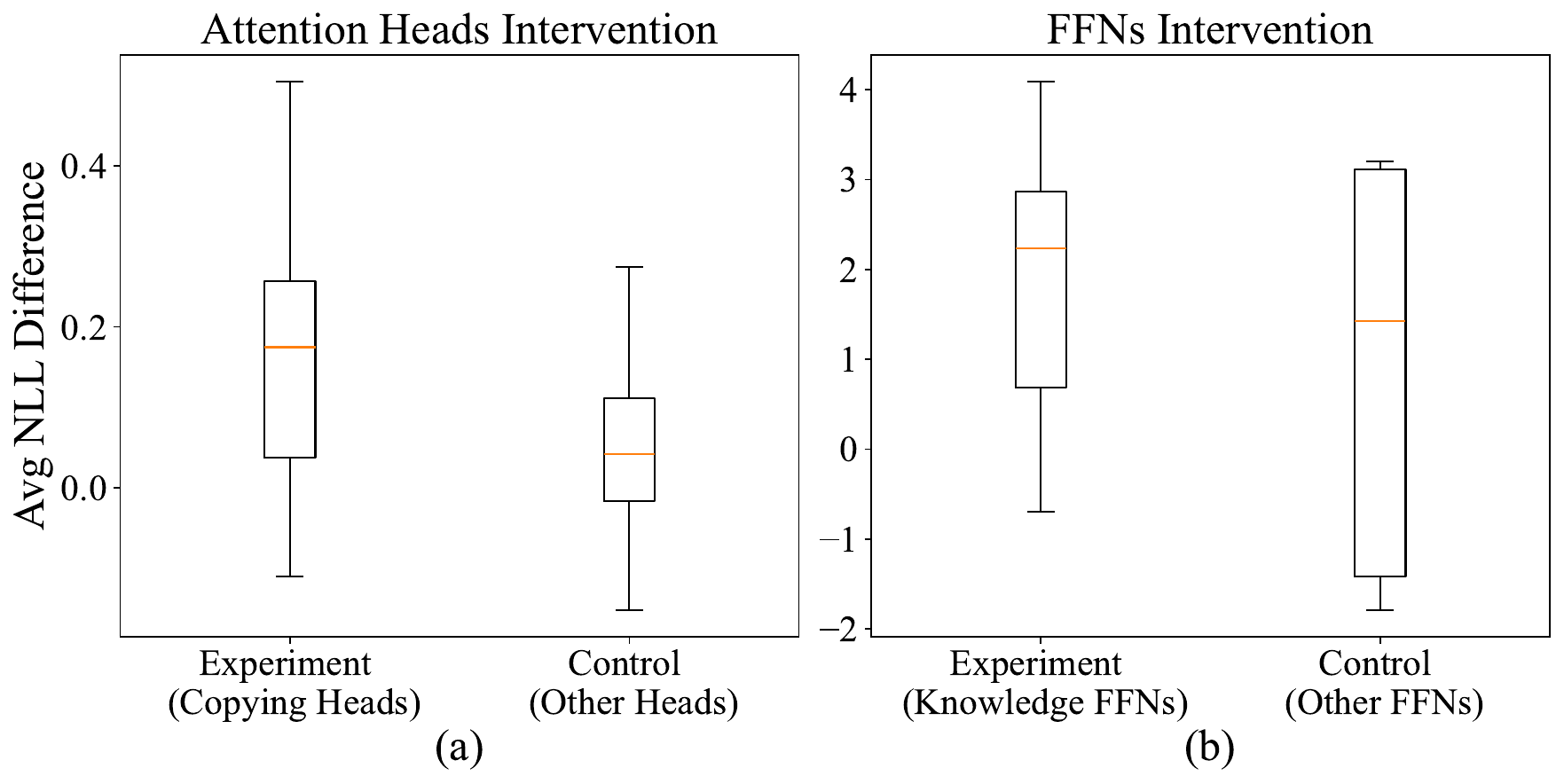}
    \end{minipage}%
    \begin{minipage}[t]{0.48\textwidth}
        \centering
        \includegraphics[width=\textwidth]{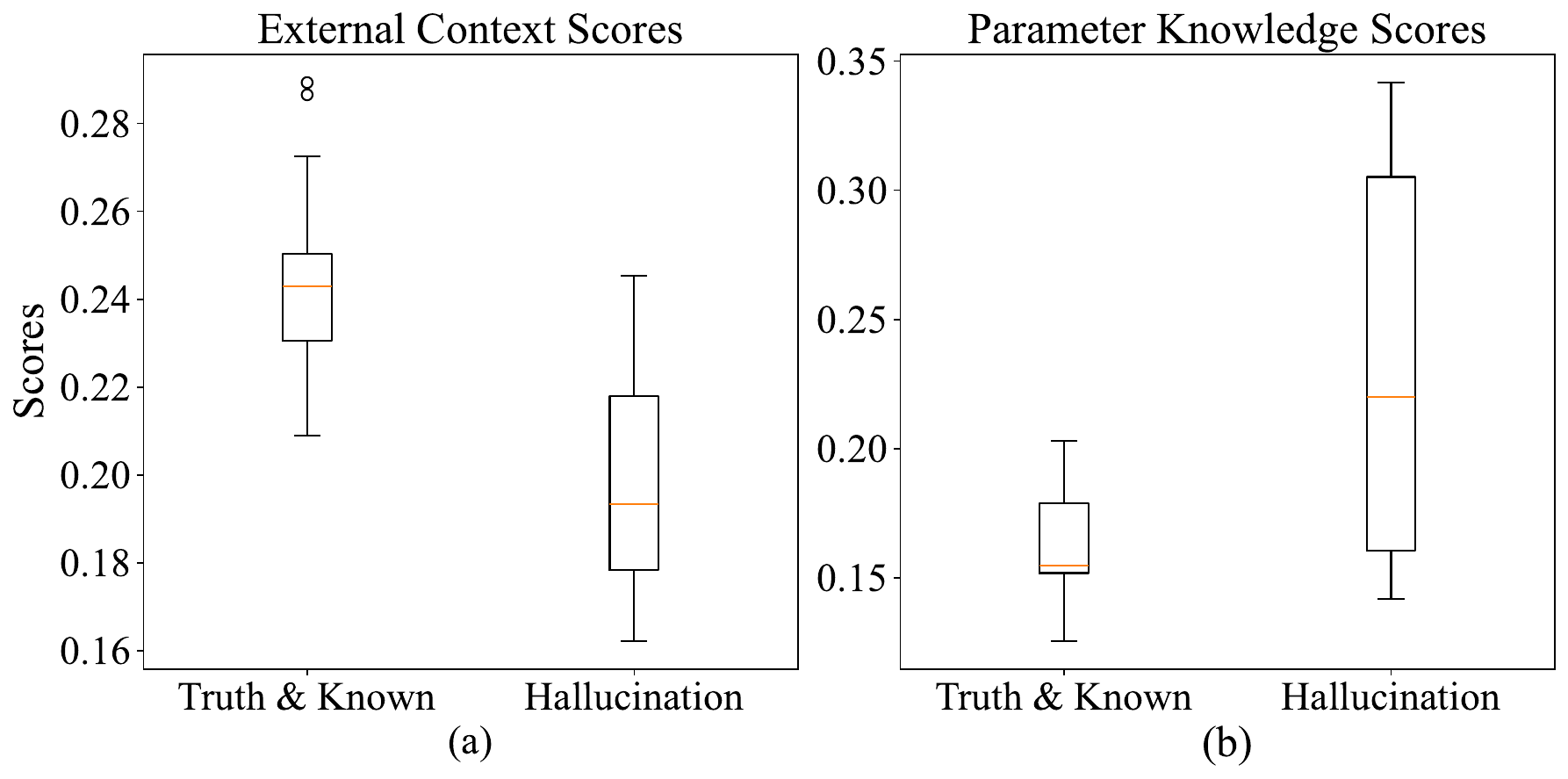}
    \end{minipage}
    \caption{(\emph{Left}) Intervention Result for Attention Heads and FFNs. (\emph{Right}) External Context Scores and Parametric Knowledge Scores (scaled by $1e^{5}$) comparing Truth \& Known (where LLM knows the truthful answer) and Hallucination (where LLM is unknown about the answer and hallucinated).}
    \label{fig:int_know_unknow}
\end{figure}

\hypertarget{rq2}{{\emph{\textbf{RQ2:} Can the relationship identified in RQ1 be validated from a causal perspective?}
\label{sec:rq2}}}

To validate the causal relationship between Copying Heads, Knowledge FFN modules, and RAG hallucinations identified in~\Secref{sec:rq1}, we employ \textbf{Causal Intervention}~\citep{ferrando2024primer} by intervening on attention heads and FFNs, where we applied noise to the attention scores and amplified the contributions of FFN modules to the residual stream (as shown in~\Figref{fig:method} (b) and (c)). We compare the Negative Log-Likelihood Loss (NLL)~\citep{pytorch_nllloss} difference for the experimental group (Copying Heads/Knowledge FFNs) and the control group (Other heads/FFNs) on truthful dataset \(\mathcal{D}^{T}\). The detailed intervention procedures are provided in the Appendix~\ref{sec:inter}. 


\textbf{Result:} As shown in~\Figref{fig:int_know_unknow} (\emph{Left}), the experimental group's impact on NLL difference was significantly greater than that of the control group for both attention heads and FFN modules. These results, combined with findings from~\Secref{sec:truth}, validate parametric knowledge added by Knowledge FFNs and the ability of Copying Heads to retrieve relevant external knowledge and LLM effectively utilizes this information during generation, have a significant causal relationship with the RAG hallucinations.

\begin{tcolorbox}[colback=gray!20, colframe=gray, coltitle=white, fonttitle=\bfseries,boxsep=0mm]
\textbf{Finding:} The occurrence of RAG hallucinations is causally related to two primary factors: \\ (1) while the Copying Heads may occasionally neglect necessary knowledge from the external context, a more prominent cause is the LLM losing the Copying Heads retrieved information during the generation process (\hyperlink{rq1-1}{RQ1-1}, \hyperlink{rq2}{RQ2}, \S~\ref{sec:dive_ecs}), and (2) the Knowledge FFNs within LLM excessively injecting parametric knowledge into the residual stream (\hyperlink{rq1-2}{RQ1-2}, \hyperlink{rq2}{RQ2}).
\end{tcolorbox}

\emph{\textbf{RQ3:} Hallucination Behavior Analysis from the Parametric Knowledge Perspective}

In this section, we focus on parametric knowledge to analyze hallucination behavior when the LLM either knows or does not know the truthful answer. We conducted a comparison experiment using the LLM-known dataset \(\widehat{\mathcal{D}}^{T}\) and the hallucination dataset \(\mathcal{D}^{H}\) (For detailed analysis, see Appendix~\ref{sec:detail_rq3}). 

\textbf{Result:} Our results in~\Figref{fig:int_know_unknow} (\emph{Right}) show that when the LLM knows the truthful answer, Copying Heads more accurately capture and utilize external knowledge, and Knowledge FFNs add less parametric knowledge to the residual stream compared to hallucination scenarios, which also supports by~\citep{wadhwa2024rags}. The results support leveraging our \textbf{Finding} to detect RAG hallucination.

\section{Methods}
\label{sec:methods}
Building on our empirical findings (\S~\ref{sec:empirical}), we propose \textbf{ReDeEP} (Regressing Decoupled External Context and Parametric Knowledge) to detect hallucinations in LLM-based retrieval-augmented generation (\S~\ref{sec:token_det}, \S~\ref{sec:chunk_det}), and \textbf{AARF} (Add Attention Reduce FFN) to mitigate hallucinations by reweighting the contributions of Knowledge FFNs and Copying Heads to the residual stream (\S~\ref{sec:truth}).

\subsection{Token-level Hallucination Detection --- ReDeEP (token)}
\label{sec:token_det}

Our empirical study identified RAG hallucinations as stemming from insufficient utilization of external context by Copying Heads (set \(\mathcal{A}\)) and excessive reliance on parametric knowledge by Knowledge FFNs (set \(\mathcal{F}\)). To address the confounding issues shown in~\Figref{fig:related_work}, we developed a multivariate analysis approach that regresses decoupled External Context Score and Parametric Knowledge Score to predict hallucinations~\citep{kahlert2017control}. For a response \(\mathbf{r}\), the hallucination score \({\mathcal{H}}_{t}\) is:

\[
{\mathcal{H}}_{t}(\mathbf{r}) =  \frac{1}{|\mathbf{r}|}\sum_{t \in \mathbf{r}}{\mathcal{H}}_{t}(t), \quad {\mathcal{H}}_{t}(t) = \sum_{l \in \mathcal{F}} \alpha \cdot\mathcal{P}^{l}_{t} - \sum_{l,h \in \mathcal{A}} \beta\cdot  \mathcal{E}^{l,h}_{t},
\]
where \(\alpha\) and \(\beta\) are regression coefficients for external context and parametric knowledge with \(\alpha, \beta > 0\), and this linear regression leverages the high Pearson correlation identified in \S~\ref{sec:empirical}.

\subsection{Chunk-level Hallucination Detection --- ReDeEP (chunk)}
\label{sec:chunk_det}
 As the \textit{Token-level Hallucination Detection} computes scores for each token, it is computationally expensive and lacks full contextual consideration. To improve efficiency and accuracy, we propose \textit{Chunk-level Hallucination Detection} as a more suitable method for RAG hallucination detection. Our approach is inspired by the common chunking operation in RAG~\cite{fan2024survey,finardi2024chronicles}, where the retrieved context \(\mathbf{c}\) and the response \(\mathbf{r}\) are divided into manageable segments \(\left\langle\boldsymbol{\tilde{c}}_i\right\rangle_{i=1}^{N}\) and \(\left\langle\boldsymbol{\tilde{r}}_j\right\rangle_{j=1}^{M}\). 
For the chunk-level external context score \(\hat{\mathcal{E}}^{l,h}\), we first calculate chunk-level attention weights \(W_{i,j}^{l,h} = \operatorname{Mean-Pooling}\left(A_{\mathbf{\tilde{c}}_i, \mathbf{\tilde{r}}_j}^{l,h}\right)\),  where \(A\) is the original token-level attention weight matrix, then determine the highest attention chunk pairs \((\mathbf{\tilde{c}}, \mathbf{\tilde{r}})\). Using an embedding model ($\operatorname{emb}$), we compute the external context score for each chunk as follows:

\[
\tilde{\mathcal{E}}^{l,h}_{\mathbf{r}} =  \frac{1}{M}\sum_{\tilde{\mathbf{r}} \in \mathbf{r}} \tilde{\mathcal{E}}^{l,h}_{\tilde{\mathbf{r}}}, \quad \tilde{\mathcal{E}}_{\mathbf{\tilde{r}}}^{l,h} = \frac{\operatorname{emb}(\mathbf{\tilde{r}}) \cdot \operatorname{emb}(\mathbf{\tilde{c}})}{\|\operatorname{emb}(\mathbf{\tilde{r}})\| \|\operatorname{emb}(\mathbf{\tilde{c}})\|}.
\]

For the chunk-level parametric knowledge score \(\tilde{\mathcal{P}}^{l}\), we sum the token-level parametric knowledge scores for each chunk:

\[
\tilde{\mathcal{P}}^{l}_{\mathbf{r}} =  \frac{1}{M}\sum_{\tilde{\mathbf{r}} \in \mathbf{r}} \tilde{\mathcal{P}}^{l}_{\tilde{\mathbf{r}}}, \quad  \tilde{\mathcal{P}}^{l}_{\mathbf{\tilde{r}}} =  \frac{1}{|\mathbf{\tilde{r}}|}\sum_{t \in \mathbf{\tilde{r}}} \mathcal{P}^{l}_{t}.
\]

Finally, the Chunk-level Hallucination Detection score \({\mathcal{H}}_{c}(\mathbf{r})\) is defined as:

\[
{\mathcal{H}}_{c}(\mathbf{r}) = \sum_{l \in \mathcal{F}} \alpha \cdot \tilde{\mathcal{P}}^{l}_{\mathbf{r}} - \sum_{l,h \in \mathcal{A}} \beta \cdot \tilde{\mathcal{E}}^{l,h}_{\mathbf{r}}.
\]

\subsection{Truthful RAG Generation  --- AARF}
\label{sec:truth}

Building on the above methods and the analysis in Appendix ~\ref{sec:dive_ecs}, we propose \textbf{Add Attention Reduce FFN (AARF)} to reduce RAG hallucinations by intervening on attention heads and FFN modules without updating model parameters. AARF operates in two stages: (1) token-level hallucination detection and (2) reweighting the contributions of attention heads and FFN modules to the residual stream. 

During the generation of token \(t_n\), we compute the hallucination score \({\mathcal{H}}_{t}(t_n)\). If \({\mathcal{H}}_{t}(t_n) \leq \tau\), we proceed with the normal output computation \(f(\mathbf{x})\) (see \Eqref{eq:final_output}). If \(\mathcal{H}_{t}(t_n) > \tau\), we adjust the weights of Copying Heads \(\mathcal{A}\) and Knowledge FFN modules \(\mathcal{F}\), shifting focus toward external context and reducing reliance on parametric knowledge:

\[
f(\mathbf{x}) = \sum_{l=1}^L \sum_{h=1}^H \widehat{\operatorname{Attn}}^{l, h}\left(\boldsymbol{X}_{\leq n}^{l-1}\right) \boldsymbol{W}_U + \sum_{l=1}^L \widehat{\operatorname{FFN}}^l\left(\boldsymbol{x}_n^{\mathrm{mid}, l}\right) \boldsymbol{W}_U + \boldsymbol{x}_n \boldsymbol{W}_U,
\]

\[
\widehat{\operatorname{Attn}}^{l, h}(\cdot) = 
\begin{cases}
\alpha_2 \cdot \operatorname{Attn}^{l, h}\left(\boldsymbol{X}_{\leq n}^{l-1}\right), & \text{if } (l, h) \in \mathcal{A}, \\
\operatorname{Attn}^{l, h}\left(\boldsymbol{X}_{\leq n}^{l-1}\right), & \text{otherwise}
\end{cases},
\quad
\widehat{\operatorname{FFN}}^l(\cdot) = 
\begin{cases}
\beta_2 \cdot \operatorname{FFN}^l\left(\boldsymbol{x}_n^{\mathrm{mid}, l}\right), & \text{if } l \in \mathcal{F}, \\
\operatorname{FFN}^l\left(\boldsymbol{x}_n^{\mathrm{mid}, l}\right), & \text{otherwise}.
\end{cases}
\]

Here, \(\alpha_2\) is a constant greater than 1 for amplifying attention head contributions, and \(\beta_2\) is a constant between (0, 1) for reducing FFN contributions.

\section{Experiments}
\label{sec:exp}
\begin{table}[t]
    \centering
    \resizebox{0.96\linewidth}{!}{
        \renewcommand\arraystretch{1.1}
    \centering
    \setlength{\tabcolsep}{1.5mm}
    \begin{tabular}{l|c|c|cccc|cccc}
    \toprule 
        \multirow{2}{*}{\textbf{LLMs}} & \multirow{2}{*}{\textbf{Categories}} & \multirow{2}{*}{\textbf{Models}}& \multicolumn{4}{c}{\textbf{RAGTruth}} & \multicolumn{4}{c}{\textbf{Dolly (AC)}}\\ \cmidrule(lr){4-7} \cmidrule(lr){8-11}
        ~ & ~ & ~ & \textbf{AUC} & \textbf{PCC} & \textbf{Rec.} & {$\mathbf{F_1}$} & \textbf{AUC} & \textbf{PCC} & \textbf{Rec.} & {$\mathbf{F_1}$} \\ \midrule 
        \multirow{18}{*}{\textbf{LLaMA2-7B}} & \multirow{5}{*}{\textbf{MPE}} & SelfCheckGPT & -- & -- & 0.3584 & 0.4642 & -- & -- & 0.1897 & 0.3188 \\ 
        ~ & ~ & Perplexity & 0.5091 & -0.0027 & 0.5190 & 0.6749 & 0.6825 & 0.2728 & 0.7719 & 0.7097 \\ 
        ~ & ~ & LN-Entropy & 0.5912 & 0.1262 & 0.5383 & 0.6655 & 0.7001 & 0.2904 & 0.7368 & 0.6772 \\ 
        ~ & ~ & Energy & 0.5619 & 0.1119 & 0.5057 & 0.6657 & 0.6074 & 0.2179 & 0.6316 & 0.6261 \\ 
        ~ & ~ & Focus & 0.6233 & 0.2100 & 0.5309 & 0.6622 & 0.6783 & 0.3174 & 0.5593 & 0.6534 \\ \cmidrule(lr){2-11}
        ~ & \multirow{7}{*}{\textbf{ECP}} & Prompt & -- & -- & 0.7200 & 0.6720 & -- & -- & 0.3965 & 0.5476 \\ 
        ~ & ~ & LMvLM & -- & -- & 0.7389 & 0.6473 & -- & -- & 0.7759 & 0.7200 \\ 
        ~ & ~ & ChainPoll & 0.6738 & 0.3563 & \underline{0.7832} & \underline{0.7066} & 0.6593 & 0.3502 & 0.4138 & 0.5581 \\ 
        ~ & ~ & RAGAS & 0.7290 & 0.3865 & 0.6327 & 0.6667 & 0.6648 & 0.2877 & 0.5345 & 0.6392 \\ 
        ~ & ~ & Trulens & 0.6510 & 0.1941 & 0.6814 & 0.6567 & \underline{0.7110} & 0.3198 & 0.5517 & 0.6667 \\ 
        ~ & ~ & RefCheck & 0.6912 & 0.2098 & 0.6280 & 0.6736 & 0.6494 & 0.2494 & 0.3966 & 0.5412 \\ 
        ~ & ~ & P(True) & 0.7093 & 0.2360 & 0.5194 & 0.5313 & 0.6011 & 0.1987 & 0.6350 & 0.6509 \\ \cmidrule(lr){2-11}
        ~ & \multirow{4}{*}{\textbf{PCE}}  & EigenScore & 0.6045 & 0.1559 & 0.7469 & 0.6682 & 0.6786 & 0.2428 & 0.7500 & 0.7241 \\ 
        ~ & ~ & SEP & 0.7143 & 0.3355 & 0.7477 & 0.6627 & 0.6067 & 0.2605 & 0.6216 & 0.7023 \\ 
        ~ & ~ & SAPLMA & 0.7037 & 0.3188 & 0.5091 & 0.6726 & 0.5365 & 0.0179 & 0.5714 & 0.7179 \\ 
        ~ & ~ & ITI & 0.7161 & 0.3932 & 0.5416 & 0.6745 & 0.5492 & 0.0442 & 0.5816 & 0.6281 \\ \cmidrule(lr){2-11}
        ~ & \multirow{2}{*}{\textbf{Ours}} & \textbf{ReDeEP(token)} & \underline{0.7325} & \underline{0.3979} & 0.6770 & 0.6986 & 0.6884 & \underline{0.3266} & \underline{0.8070} & \underline{0.7244} \\ 
        ~ & ~ & \textbf{ReDeEP(chunk)} & \bf 0.7458 & \bf 0.4203 & \bf 0.8097 &  \bf 0.7190 & \bf 0.7949 & \bf 0.5136 & \bf 0.8245 & \bf 0.7833 \\ \midrule \midrule
        \multirow{18}{*}{\textbf{LLaMA2-13B}} & \multirow{5}{*}{\textbf{MPE}} & SelfCheckGPT & -- & -- & 0.3584 & 0.4642 & -- & -- & 0.1897 & 0.3188 \\ 
        ~ & ~ & Perplexity & 0.5091 & -0.0027 & 0.5190 & 0.6749 & 0.6825 & 0.2728 & 0.7719 & 0.7097 \\ 
        ~ & ~ & LN-Entropy & 0.5912 & 0.1262 & 0.5383 & 0.6655 & 0.7001 & 0.2904 & 0.7368 & 0.6772 \\ 
        ~ & ~ & Energy & 0.5619 & 0.1119 & 0.5057 & 0.6657 & 0.6074 & 0.2179 & 0.6316 & 0.6261 \\ 
        ~ & ~ & Focus & 0.7888 & 0.4444 & 0.6173 & 0.6977 & 0.7067 & 0.1643 & 0.7333 & 0.6168 \\ \cmidrule(lr){2-11}
        ~ & \multirow{7}{*}{\textbf{ECP}} & Prompt & -- & -- & 0.7000 & 0.6899 & -- & -- & 0.4182 & 0.5823 \\ 
        ~ & ~ & LMvLM & -- & -- & \bf 0.8357 & 0.6553 & -- & -- & 0.7273 & 0.6838 \\ 
        ~ & ~ & ChainPoll & 0.7414 & 0.4820 & \underline{0.7874} & 0.7342 & 0.7070 & \underline{0.4758} & 0.4364 & 0.6000 \\ 
        ~ & ~ & RAGAS & 0.7541 & 0.4249 & 0.6763 & 0.6747 & 0.6412 & 0.2840 & 0.4182 & 0.5476 \\ 
        ~ & ~ & Trulens & 0.7073 & 0.2791 & 0.7729 & 0.6867 & 0.6521 & 0.2565 & 0.3818 & 0.4941 \\ 
        ~ & ~ & RefCheck & 0.7857 & 0.4104 & 0.6800 & 0.7023 & 0.6626 & 0.2869 & 0.2545 & 0.3944 \\ 
        ~ & ~ & P(True) & 0.7998 & 0.3493 & 0.5980 & 0.7032 & 0.6396 & 0.2009 & 0.6180 & 0.5739 \\ \cmidrule(lr){2-11}
        ~ & \multirow{4}{*}{\textbf{PCE}} & EigenScore & 0.6640 & 0.2672 & 0.6715 & 0.6637 & 0.7214 & 0.2948 & \underline{0.8181} & \underline{0.7200} \\ 
        ~ & ~ & SEP & 0.8089 & 0.5276 & 0.6580 & 0.7159 & 0.7098 & 0.2823 & 0.6545 & 0.6923 \\ 
        ~ & ~ & SAPLMA & 0.8029 & 0.3956 & 0.5053 & 0.6529 & 0.6053 & 0.2006 & 0.6000 & 0.6923 \\ 
        ~ & ~ & ITI & 0.8051 & 0.4771 & 0.5519 & 0.6838 & 0.5511 & 0.0646 & 0.5385 & 0.6712 \\ \cmidrule(lr){2-11}
        ~ &  \multirow{2}{*}{\textbf{Ours}} & \textbf{ReDeEP(token)} & \underline{0.8181} & \underline{0.5478} & 0.7440 & \underline{0.7494} & \underline{0.7226} & 0.3776 & \underline{0.8148} & 0.7154 \\ 
        ~ & ~ & \textbf{ReDeEP(chunk)} & \bf 0.8244 & \bf 0.5566 & 0.7198 & \bf 0.7587 & \bf 0.8420 & \bf 0.5902 & \bf 0.8518 & \bf 0.7603 \\ \midrule \midrule
         \multirow{18}{*}{\textbf{LLaMA3-8B}} & \multirow{5}{*}{\textbf{MPE}} & SelfCheckGPT & -- & --  & 0.4111 & 0.5111 & -- & -- &  0.2195 & 0.3600 \\ 
        ~ & ~ & Perplexity & 0.6235 & 0.2100 &  0.6537 & 0.6778 & 0.5924 & 0.1095  & 0.3902 & 0.4571 \\ 
        ~ & ~ & LN-Entropy & 0.7021 & 0.3451  & 0.5596 & 0.6282 & 0.6011 & 0.1150  & 0.5365 & 0.5301 \\ 
        ~ & ~ & Energy & 0.5959 & 0.1393  & 0.5514 & 0.6720 & 0.5014 & -0.0678 &  0.4047 & 0.5440 \\ 
        ~ & ~ & Focus & 0.6378 & 0.3079 & 0.6688 & 0.6879 & 0.6177 & 0.1266 &  0.6918 & 0.6874 \\ \cmidrule(lr){2-11}
        ~ & \multirow{7}{*}{\textbf{ECP}} & Prompt & -- & -- & 0.4403 & 0.5691 & -- & -- &  0.3902 & 0.5000 \\ 
        ~ & ~ & LMvLM & -- & -- &  0.5109 & 0.6986 & -- & --  & 0.6341 & 0.5361 \\ 
        ~ & ~ & ChainPoll & 0.6687 & 0.3693  & 0.4486 & 0.5813 & 0.6114 & 0.2691 &  0.3415 & 0.4516 \\ 
        ~ & ~ & RAGAS & 0.6776 & 0.2349  & 0.3909 & 0.5094 & 0.6870 & \underline{0.3628}  & 0.8000 & 0.5246 \\ 
        ~ & ~ & Trulens & 0.6464 & 0.1326  & 0.3909 & 0.5053 & \underline{0.7040} & 0.3352 &  0.3659 & 0.5172 \\ 
        ~ & ~ & RefCheck & 0.6014 & 0.0426  & 0.3580 & 0.4628 & 0.5260 & -0.0089  & 0.1951 & 0.2759 \\ 
        ~ & ~ & P(True) & 0.6323 & 0.2189  & 0.7083 & 0.6835 & 0.6871 & 0.3472  & 0.5707 & 0.6573 \\ \cmidrule(lr){2-11}
        ~ &\multirow{4}{*}{\textbf{PCE}}& EigenScore & 0.6497 & 0.2120  & 0.7078 & 0.6745 & 0.6612 & 0.2065  & 0.7142 & 0.5952 \\ 
        ~ & ~ & SEP & 0.7004 & 0.3713  & 0.7333 & 0.6915 & 0.5159 & 0.0639  & 0.6829 & 0.5385 \\ 
        ~ & ~ & SAPLMA & 0.7092 & 0.4054 &  0.5432 & 0.6718 & 0.5019 & -0.0327  & 0.4040 & 0.5714 \\ 
        ~ & ~ & ITI & 0.6534 & 0.3404  & 0.6850 & 0.6933 & 0.5011 & 0.0024  & 0.3091 & 0.4250 \\ \cmidrule(lr){2-11}
        ~ &  \multirow{2}{*}{\textbf{Ours}} & \textbf{ReDeEP(token)} & \bf 0.7522 & \bf 0.4493 &   \bf 0.7984 & \bf 0.7132 & 0.6701 & 0.2421 &  \underline{0.8293} & \underline{0.6901} \\ 
        ~ & ~ & \textbf{ReDeEP(chunk)} & \underline{0.7285} & \underline{0.3964}  & \underline{0.7819} & \underline{0.6947} & \bf 0.7354 & \bf 0.3652  & \bf 0.8392 & \bf 0.7100 \\ \bottomrule
    \end{tabular}
     \caption{Performance comparisons between ReDeEP and the baselines. The boldface represents the best performance, and the underline represents the second-best. }
    \label{tab:main_results}}
\end{table}

\subsection{Settings}
\label{sec:setting}
\textbf{Data:} We evaluate ReDePE and AARF on two public RAG hallucination datasets.
\textbf{RAGTruth} is the first high-quality, manually annotated RAG hallucination dataset. The data includes three RAG task types: Question Answering (QA), Data-to-Text Writing, and News Summarization.
\textbf{Dolly (AC)} is a dataset with Accurate Context obtained from~\citep{hu2024refchecker}, including tasks such as text summarization, closed-QA, and information extraction. More details of the data are in Appendix~\ref{sec:data_details}.

\textbf{Baselines:} We conduct experiments on three variants of LLaMA, including LLaMA2-7B-Chat, LLaMA2-13B-Chat, and LLaMA3-8B-Chat. For hallucination detection methods, we follow the classification of existing methods as shown in~\Figref{fig:related_work}. We use (1) Parametric Confounded by External Methods (\textbf{PCE}), {(2) External Confounded by Parametric Methods} (\textbf{ECP}), and {(3) Mixed Parametric and External Methods} (\textbf{MPE}). For detailed baselines information, see Appendix~\ref{sec:baseline_de}. We used AUC, Pearson Correlation Coefficient (PCC), Accuracy (Acc.), Recall (Rec.), and $\text{F}_{1}$ as evaluation metrics for detection accuracy. Implementation details are provided in Appendix~\ref{sec:imp}.




\subsection{Experiments}
\textbf{RAG Hallucination Detection:} As shown in Table~\ref{tab:main_results}, ReDeEP consistently improves performance across two datasets, various backbone methods, and different metrics, validating its effectiveness in detecting RAG hallucinations. ReDeEP outperforms MPE methods, demonstrating that mechanistic interpretability effectively decouples the LLM's utilization of external context and parametric knowledge, enabling more accurate detection of RAG hallucinations.
Additionally, ReDeEP surpasses both ECP and PCE methods by incorporating both the External Context Score and Parametric Knowledge Score as covariates in a multivariate regression approach, effectively addressing the confounding problem. 
ReDeEP(chunk) generally outperforms ReDeEP(token) in most metrics, suggesting that chunk-level processing better preserves semantic integrity and improves detection performance. Further support for ReDeEP's effectiveness is provided by the ablation study in Appendix~\ref{sec:ablation}, while the efficiency analysis in Appendix~\ref{sec:eff} confirms that ReDeEP achieves comparable time efficiency to the most efficient baselines.


\begin{figure}[t]
    \centering
    \begin{subfigure}{0.5\linewidth}
        \centering
        \includegraphics[width=\linewidth]{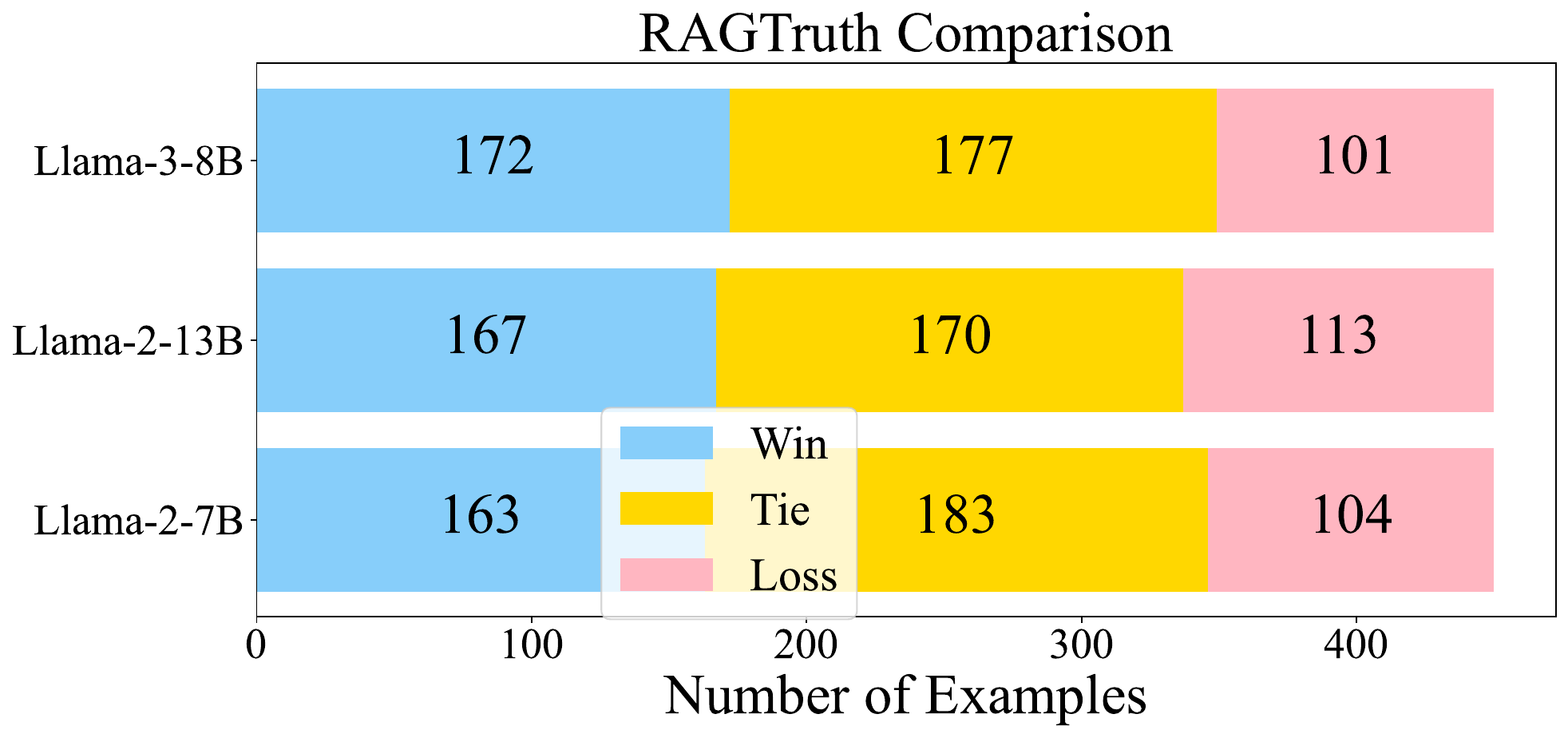}
        \label{fig:dolly-truth}
    \end{subfigure}%
    \begin{subfigure}{0.5\linewidth}
        \centering
        \includegraphics[width=\linewidth]{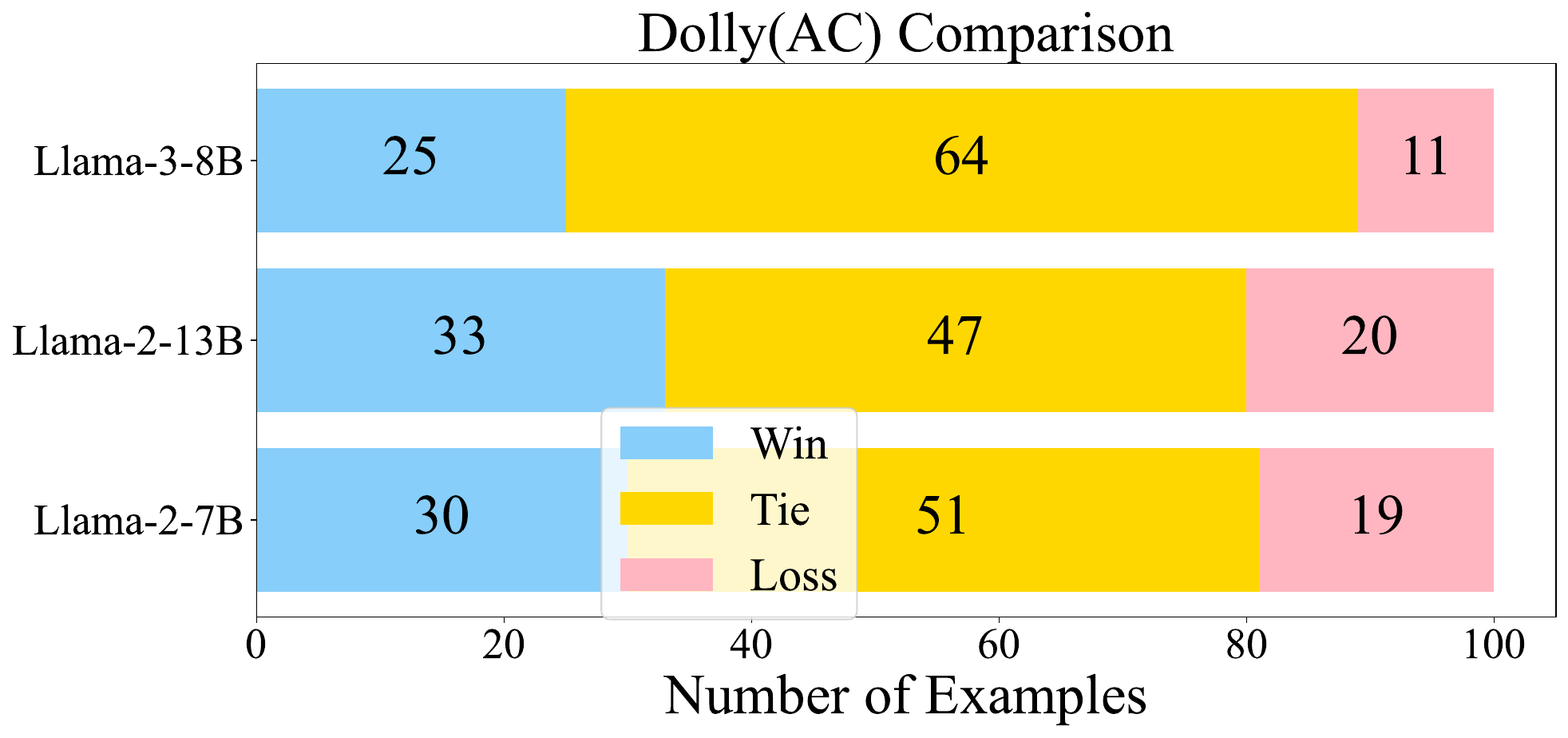}
        \label{fig:ragtruth-truth}
    \end{subfigure}
    \caption{Comparison between LLMs+AARF vs LLMs judged by GPT-4o.}
    \label{fig:reduce_hallucination}
\end{figure}
\textbf{Truthful RAG Generation:}
We evaluated our hallucination reduction method AARF on both the RAGTruth and Dolly (AC) datasets, using GPT-4-o for automatic evaluation to assess the truthfulness (Prompt details can be found in Appendix~\ref{sec:prompt_truthful_eval}). Pairwise comparisons rated by GPT-4-o are shown in Figure~\ref{fig:reduce_hallucination}, demonstrating that AARF can reduce hallucinations to a certain extent compared to the baseline model. These results validate the effectiveness of our intervention experiments and confirm the findings presented in \textbf{RQ2} of Section~\ref{sec:rq1}.

\section{Conclusion}

Detecting RAG hallucinations is critical for enhancing the security and reliability of RAG systems. In this work, we introduced ReDeEP, a novel method that detects RAG hallucinations by analyzing LLMs' utilization of parametric knowledge and external context. Our empirical study shows that hallucinations arise from insufficient utilization of external context by Copying Heads and over-reliance on parametric knowledge by Knowledge FFN modules. These insights also guided the development of interventions to reduce hallucinations without updating model parameters. ReDeEP demonstrates significant performance improvements across the LLaMA family and RAG hallucination benchmarks, outperforming existing detection methods.



\bibliography{iclr2025_conference}
\bibliographystyle{iclr2025_conference}

\appendix
\section{Full Background on Attention Heads, FFNs, and Logit Lens}
\label{sec:back_details}
The theoretical foundation of our work is grounded in research on mechanistic interpretability~\citep{ferrando2024primer,nostalgebraist2020,meng2022locating,transformercircuits2021}, which seeks to explain the internal processes of language models (LMs) by interpreting how individual model components contribute to the final prediction. In this study, we focus on the transformer decoder-only architecture (also known as GPT-like) due to its widespread success and popularity~\citep{achiam2023gpt,dubey2024llama}. A decoder-only model $f$ consists of $L$ layers and operates on a sequence of embeddings $\mathbf{x} = \left\langle \boldsymbol{x}_1, \boldsymbol{x}_2, \ldots, \boldsymbol{x}_n \right\rangle$, which represent the tokens $\mathbf{t} = \left\langle t_1, t_2, \ldots, t_n \right\rangle$. Each embedding $\boldsymbol{x} \in \mathbb{R}^d$ is a row vector corresponding to a row of the embedding matrix $\boldsymbol{W}_E \in \mathbb{R}^{|\mathcal{V}| \times d}$, where $\mathcal{V}$ denotes the model's vocabulary. The sequence $\mathbf{x}$ is represented as a matrix $\boldsymbol{X}^{0} \in \mathbb{R}^{n \times d}$ with the embeddings stacked as rows.

We interpret Transformers through the perspective of the residual stream \citep{transformercircuits2021}. Due to the residual connections in Transformers, each layer \( l \) takes a hidden state \(\boldsymbol{X}^{l-1}\) as input and adds information obtained from its \textit{Attention Heads} and Feed-Forward Networks (\textit{FFNs}) to the hidden state via the residual connection. In this context, the hidden state acts as a residual stream passed through the layers, with each attention and FFN contributing to the final prediction by adding information to the residual stream, resulting in the \textit{Residual Stream States}. The final layer's residual stream state is then projected into the vocabulary space using the \textit{Unembedding Matrix} \(\boldsymbol{W}_U \in \mathbb{R}^{d \times |\mathcal{V}|}\) and normalized via the softmax function to produce a probability distribution over the vocabulary, from which a new token is sampled.

The background knowledge for interpreting the contributions of each FFN and attention head to the model’s prediction is outlined as follows:

\textbf{Attention Heads:} Attention is crucial in Transformers for contextualizing token representations across layers. Each attention head selectively attends to previous positions, gathers information, and updates the current residual stream~\citep{ferrando2024information,clark-etal-2019-bert,wu2024retrieval}. The output of an attention layer is the sum of its attention heads. For each attention head:

\begin{equation}
    \operatorname{Attn}^{l, h}\left(\boldsymbol{X}_{\leq i}^{l-1}\right) = \sum_{j \leq i} a_{i, j}^{l, h} \boldsymbol{x}_j^{l-1} \boldsymbol{W}_V^{l, h} \boldsymbol{W}_O^{l, h} = \sum_{j \leq i} a_{i, j}^{l, h} \boldsymbol{x}_j^{l-1} \boldsymbol{W}_{OV}^{l, h}
\end{equation}

where the learnable weight matrices \(\boldsymbol{W}_V^{l, h} \in \mathbb{R}^{d \times d_h}\) and \(\boldsymbol{W}_O^{l, h} \in \mathbb{R}^{d_h \times d}\) are combined into the OV matrix \(\boldsymbol{W}_{OV}^{l, h} = \boldsymbol{W}_V^{l, h} \boldsymbol{W}_O^{l, h} \in \mathbb{R}^{d \times d}\), also referred to as the OV \textit{(output-value)} circuit~\citep{kobayashi2021incorporating}. The attention weights for the current query \(i\) to the previous tokens are computed as:

\begin{equation}
\boldsymbol{a}_i^{l, h} = \operatorname{softmax}\left(\frac{\boldsymbol{x}_i^{l-1} \boldsymbol{W}_Q^{l, h} \left(\boldsymbol{X}_{\leq i}^{l-1} \boldsymbol{W}_K^{l, h}\right)^{\top}}{\sqrt{d_k}}\right) = \operatorname{softmax}\left(\frac{\boldsymbol{x}_i^{l-1} \boldsymbol{W}_{QK}^h \boldsymbol{X}_{\leq i}^{l-1\top}}{\sqrt{d_k}}\right)
\label{eq:attention_weight}
\end{equation}

where \(\boldsymbol{W}_Q^{l, h} \in \mathbb{R}^{d \times d_h}\) and \(\boldsymbol{W}_K^{l, h} \in \mathbb{R}^{d \times d_h}\) combine to form the QK (query-key) circuit~\citep{transformercircuits2021}, \(\boldsymbol{W}_{QK}^h = \boldsymbol{W}_Q^{h} \boldsymbol{W}_K^{h\top} \in \mathbb{R}^{d \times d}\). The QK circuit computes the attention weights, determining the positions that should be attended, while the OV circuit transfers and transforms the information from the attended positions into the current residual stream. The attention block output is the sum of individual attention heads, which is then added back into the residual stream:

\begin{equation}
\boldsymbol{x}_i^{\mathrm{mid}, l} = \boldsymbol{x}_i^{l-1} + \sum_{h=1}^H \operatorname{Attn}^{l, h}\left(\boldsymbol{X}_{\leq i}^{l-1}\right)
\label{eq:x_mid}
\end{equation}

Previous research has shown that the primary role of attention layers in LLMs is implementing algorithms~\citep{olsson2022context,ferrando2024primer}. For instance, several attention heads in Transformer LMs have OV matrices exhibiting copying behavior; we refer to these heads as \textit{\textbf{Copying Heads}}. \citet{transformercircuits2021} propose using the number of positive real eigenvalues of the full OV circuit matrix \(\boldsymbol{W}_E \boldsymbol{W}_{OV} \boldsymbol{W}_U\) as a summary statistic for detecting Copying Heads. Positive eigenvalues indicate that a linear combination of tokens contributes to an increase in the linear combination of logits of the same tokens.

\textbf{FFN:} Research has shown that the functionality of FFN layers lies in storing knowledge~\citep{geva2021transformer}. Transformer FFN layers can be represented as a linear combination of vectors. Specifically, for an input vector \(\boldsymbol{x}_i^{\mathrm{mid}, l} \in \mathbb{R}^d\) drawn from the residual stream states, with FFN parameter matrices \(\mathbf{K}^{l}, \mathbf{V}^{l} \in \mathbb{R}^{d_m \times d}\), the FFN output can be expressed as:

\[
\mathrm{FFN}^{l}\left(\boldsymbol{x}^{\text{mid}, l}_{i}\right) = g\left(\boldsymbol{x}_i^{\mathrm{mid}, l} (\mathbf{K}^{l})^{T}\right) \mathbf{V}^{l} = \sum_{i=1}^{d_m} f\left(\boldsymbol{x}_i^{\mathrm{mid}, l} \cdot \boldsymbol{k}_i^{l}\right) \boldsymbol{v}_i^{l} = \sum_{i=1}^{d_m} m_i^{l} \boldsymbol{v}_i^{l}
\]

\begin{equation}
\boldsymbol{x}^{l}_{i} = \boldsymbol{x}^{\text{mid}, l}_{i} + \mathrm{FFN}^{l}\left(\boldsymbol{x}^{\text{mid}, l}_{i}\right)
\label{eq:x_final}
\end{equation}

where \(g\) is the activation function. Thus, the FFN layer can be viewed as a linear combination of vectors: the multiplication of \(\boldsymbol{x}_i^{\mathrm{mid}, l}\) and the key vector \(\boldsymbol{k}_i^{l}\) produces the coefficient \(m_i^{l}\), which weights the corresponding value vector \(\boldsymbol{v}_i^{l}\).

\textbf{Logit Lens:} The \(\operatorname{LogitLens}\) is a technique that decodes hidden states \(\boldsymbol{x}^{l}\) directly into the vocabulary distribution using the LayerNorm and the unembedding matrix of the LLM for interpretability~\citep{nostalgebraist2020}:

\begin{equation}
    \operatorname{LogitLens}\left(\boldsymbol{x}^{l}\right) = \operatorname{LayerNorm}(\boldsymbol{x}^{l}) \boldsymbol{W}_U
\end{equation}

This approach has been validated in various studies as an effective method for interpreting LLMs' weight matrices or hidden states~\citep{hanna2024does,zhou2024unibias,yu2023characterizing}.

The final output logits of the LLM can be expressed as:

\begin{equation}
    f(\mathbf{x}) = \left(\sum_{l=1}^L \sum_{h=1}^H \operatorname{Attn}^{l, h}\left(\boldsymbol{X}_{\leq n}^{l-1}\right) \boldsymbol{W}_U + \sum_{l=1}^L \operatorname{FFN}^l\left(\boldsymbol{x}_n^{\mathrm{mid}, l}\right) \boldsymbol{W}_U + \boldsymbol{x}_n \boldsymbol{W}_U\right)
    \label{eq:final_output}
\end{equation}

\section{Calculation of Copying Heads Score}
\label{sec:cal_copy_heads}
The traditional method for detecting Copying Heads~\citep{transformercircuits2021} involves calculating the eigenvalues of the matrix \( M = W_U W_{O V}^h W_E \) and assessing the proportion of positive eigenvalues, where $W_E$ is the Embedding Matrix, $W_{O V}$ is the OV Matrix and $W_U$ is the Unembedding Matrix.
However, the original Copying Heads identification method proposed by~\citet{transformercircuits2021} requires calculating eigenvalues of large matrices and using the ratio of positive eigenvalues to determine Copying Heads, which becomes computationally expensive for models with large hidden sizes. We propose using the trace of the matrix to estimate the ratio of positive eigenvalues, calibrated with the Gershgorin circle theorem~\citep{bell1965gershgorin}, to obtain each attention head’s copying head score \(\mathcal{C}^{l,h}\).
To simplify this, we estimate the proportion using the trace of the matrix and refine this estimation using the Gershgorin Circle Theorem~\citep{bell1965gershgorin}.

\begin{enumerate}
    \item \textbf{Trace-Based Estimation:} The trace \(\text{tr}(M)\) provides an indication of the distribution of positive and negative eigenvalues:
    \begin{itemize}
        \item A positive trace suggests more positive than negative eigenvalues.
        \item A negative trace suggests more negative than positive eigenvalues.
        \item A trace near zero suggests a balance between positive and negative eigenvalues.
    \end{itemize}

    \item \textbf{Gershgorin Circle Theorem:} To enhance the accuracy of our trace-based estimation, we employ the Gershgorin Circle Theorem, which provides an approximation of the eigenvalue distribution. For any \( n \times n \) matrix \( M = [a_{ij}] \), each eigenvalue of \( M \) lies within at least one Gershgorin disk \( D_i \):
    \[
    D_i = \{ z \in \mathbb{C} : |z - a_{ii}| \leq R_i \}
    \]
    where \( R_i = \sum_{j \neq i} |a_{ij}| \). Each disk is centered at the diagonal element \( a_{ii} \) with a radius determined by the sum of the absolute values of the off-diagonal elements in the row. This theorem helps identify the regions in the complex plane where the eigenvalues are likely to be found, allowing us to approximate their distribution without direct computation.

    \item \textbf{IQR-Based Outlier Detection for Boundary Points:}
    \begin{itemize}
        \item Collect boundary points \( z + a_{ii} \) and \( z - a_{ii} \) from each Gershgorin disk.
        \item Calculate the first (Q1) and third quartiles (Q3) of these points, then determine the IQR as \( \text{IQR} = Q3 - Q1 \)~\citep{vinutha2018detection}.
        \item Identify outliers using bounds \( Q1 - 1.5 \times \text{IQR} \) and \( Q3 + 1.5 \times \text{IQR} \), counting points outside these limits.
    \end{itemize}

    \item \textbf{Copying Head Score Calculation:} Combine rankings based on the number of detected outliers (ascending order) and the absolute value of the trace (descending order). Summing these ranks gives the Copying Head Score \(\mathcal{C}^{l,h}\), reflecting the head’s tendency to behave as a Copying Head.
\end{enumerate}

\section{Dive into the Rationale Behind the External Context Score}
\label{sec:dive_ecs}
In~\Secref{sec:metrics}, we designed the external context score to measure two aspects: (1) whether the attention heads focus on the correct external context, and (2) if attending to the correct context, whether the LLM can effectively retain and utilize this information during the generation process. In this section, we aim to explore whether a low external context score is caused by attention heads focusing on the incorrect external context or by the LLM losing the information attended by the attention heads during the generation process.

To address this, we conducted experiments on the Llama2-7B model using the RAGTruth dataset. Specifically, in our validation experiments, we selected data from RAGTruth where LLaMA2-7B-Chat exhibited hallucinations. Using LangChain, a widely-used open-source toolkit, we applied the RecursiveCharacterTextSplitter to segment the input retrieved document into different spans. We then calculated whether the attention module’s attended span (by mean pooling the attention scores and selecting the input span with the highest score) during the generation of hallucination spans could identify the hallucination spans in the response.  If the attention head successfully identified the correct span, this indicates that the attention mechanism focused on the correct external context, but the LLM did not effectively retain and utilize this information during the generation process.
This evaluation was based on GPT-4-o (from OpenAI) using the following prompt:

\begin{tcolorbox}[colback=gray!20, colframe=gray, coltitle=white, fonttitle=\bfseries]

    Prompt: \{external context + query\}

    Respond: \{response\}

    Conflict Span: \{Conflict Span\} \\
    Conflict Type: \{Conflict Type\} \\
    Reason: \{Reason\}

    Given the following context information: "\{Attend Span\}", can this support the existence of a conflict in the response? Please answer with "Yes" or "No" and give the reason on the newline.
\end{tcolorbox}
\begin{table}[ht]
\center
\begin{tabular}{c|c}
\toprule
Attention heads attend & Attention heads mis-attend \\
\midrule
77.5\% & 22.5\% \\
\bottomrule
\end{tabular}
\caption{Proportion of data where Llama2-7B attention heads attend to the correct information.}
\label{tab:attend}
\end{table}

From the results in Table~\ref{tab:attend}, we can see that a low external context score is mostly due to the LLM losing the information attended by the attention heads during the generation process. In most cases, the attention heads correctly attend to the appropriate external context. 
This phenomena may be due to the presence of some Copy suppression heads in the LLM~\citep{mcdougall2023copy}, which may incorrectly suppress the information attended by the Copying head, resulting in the LM losing the information attended by the attention heads during the generation process.
This also validates the feasibility of our proposed AARF method in reducing hallucinations by increasing the output of copying heads in the residual stream.

\section{Detailed Intervention Procedures}
\label{sec:inter}
In this section, we provide the details of the intervention experiments for \textbf{RQ2} from Section~\ref{sec:rq1}.

\textbf{Attention Heads Intervention:} As described in~\Figref{fig:method} (b), we applied noise to the attention scores \(\boldsymbol{a}_{i}^{l,h}\) of the experimental group to evaluate their impact on hallucinations. Specifically, the attention scores were sampled from a standard normal distribution:

\begin{equation}
a_{i,j}^{l,h} \sim \mathcal{N}(0, 1), \quad \tilde{\boldsymbol{a}}_{i}^{(l,h)} = \text{softmax}\left(\boldsymbol{a}^{(l,h)}\right).
\end{equation}

This approach simulates the removal of meaningful attention patterns, allowing us to assess how Copying Heads' focus on external context impacts hallucinations.

\textbf{FFN Modules Intervention:} To investigate the role of Knowledge FFNs in hallucinations, we amplified the effect of the FFN modules by increasing their contribution to the residual stream tenfold (\(k = 10\)). This intervention highlights the influence of parametric knowledge on the generation process.

\textbf{Causal Matching:} To reduce potential biases arising from the position of transformer layers or heads, we applied causal matching~\citep{stuart2010matching}. For attention heads, we matched the top 32 Copying Heads with the nearest non-experimental heads within the same layer. Similarly, we matched the top 5 Knowledge FFNs, identified as being most related to hallucinations, with the nearest FFN modules in adjacent layers. This matching process ensured that the comparison between the experimental and control groups was fair and focused on the specific roles of Copying Heads and Knowledge FFNs in hallucination generation.

\section{Detailed analysis of \textbf{RQ3:} Hallucination Behavior analysis from parametric knowledge view}
\label{sec:detail_rq3}

In the experiments (\textbf{RQ1} and \textbf{RQ2} in~\Secref{sec:rq1}), we analyzed the relationship between LLM-generated responses and the internal states of the model that lead to hallucinations. In this section, we shift our focus to parametric knowledge (since our setting assumes the external context is correct, there is no need for separate analysis from the external context perspective) to examine the two scenarios where the LLM's internal memory either knows or does not know the truthful answer to the query. This analysis aims to validate whether our previous findings about the connection between LLM hallucinations and internal states are reasonable.

\textbf{LLM Parametric Knowledge Unknown Truthful Answer} 
When the LLM's parametric knowledge does not contain the truthful response, the model must rely on the retrieved context to generate a truthful answer. In this scenario, the Knowledge FFN module may over-add parametric knowledge to the residual stream, while Copying Heads may fail to attend to the correct external context or lose the attended external information during the generation process, leading to hallucinations. This phenomenon is consistent with our earlier findings, where Copying Heads neglect external context and Knowledge FFN modules excessively add parametric knowledge to the residual stream.

\textbf{LLM Parametric Knowledge Known Truthful Answer} 
When the LLM's parametric knowledge contains the truthful answer, RAG responses are typically truthful. To verify whether the model, in this case, relies more on external knowledge and less on parametric knowledge compared to when hallucinations occur, we designed a validation experiment. Specifically, we allowed the LLM to generate responses directly on the truthful dataset \(\mathcal{D}^{T}\) without relying on retrieved documents to determine if the LLM could independently produce accurate answers. We used GPT-4-o~\citep{achiam2023gpt}, along with the original truthful response, to evaluate whether the LLM-generated answers matched the expected ones, thus assessing if the LLM's parametric knowledge could correctly answer independently (see prompt in Appendix~\ref{sec:Prompt_know}). These correct responses form the LLM-known dataset \(\widehat{\mathcal{D}}^{T}\). We then analyzed the differences in external context scores and parametric knowledge scores for the LLM across \(\widehat{\mathcal{D}}^{T}\) and \(\mathcal{D}^{H}\).

\textbf{Result:} As shown in~\Figref{fig:int_know_unknow} (\emph{Right}), when the LLM’s parametric knowledge knows the truthful answer, we observe that Copying Heads can more accurately capture external knowledge and effectively retain and utilize this information during the generation process, showing more stable performance compared to their behavior in the hallucination dataset. Although in scenarios where the LLM knows the truthful answer, the Knowledge FFN layers add less parametric knowledge to the residual stream than the hallucination dataset, this supports our earlier finding of the negative impact of excessive utilization of parametric knowledge by the Knowledge FFN module.

\section{Prompt for Evaluating Parametric Knowledge}
\label{sec:Prompt_know}

To assess whether the LLM's parametric knowledge alone could provide accurate answers independently of retrieved documents, we used the following prompt in our validation experiment. The aim was to evaluate if the LLM-generated responses on the truthful dataset \(\mathcal{D}^{T}\) matched the expected truthful answers. The prompt was designed to engage GPT-4-o~\citep{achiam2023gpt} as an evaluator to compare the LLM's responses with the ground truth.

\textbf{Prompt:}
\begin{tcolorbox}[colback=gray!20, colframe=gray, coltitle=white, fonttitle=\bfseries]
You are an AI evaluator tasked with assessing the accuracy and relevance of an AI-generated response. Here are the details:

\begin{enumerate}
    \item AI-generated response: \textit{\{LLM-Generated-Response\}}
    \item Expected response: \textit{\{Ground Truth\}}
    \item Query that prompted the response: \textit{\{Query\}}
\end{enumerate}

Evaluate if the AI-generated response accurately and comprehensively addresses the query and aligns with the expected response. If the AI-generated response aligns well with the expected response, output "yes". If it does not align, output "no". Only output "yes" or "no".
\end{tcolorbox}

\section{Details about RAG Hallucination Datasets}
\label{sec:data_details}
\textbf{RAGTruth:} The RAGTruth~\citep{niu-etal-2024-ragtruth} dataset is the first high-quality, manually annotated RAG hallucination dataset. It is divided into three task types: Question Answering (QA), Data-to-Text Writing, and News Summarization. However, the dataset does not include hallucination annotations for responses generated by Llama3-8B. To address this, we employed three annotators with graduate-level qualifications to manually evaluate the presence of hallucinations in different LLM RAG responses. Each response was carefully assessed to determine whether it contained any hallucinations based on the accuracy and relevance of the retrieved and generated content.

\textbf{Dolly (AC):} The Dolly (AC) dataset is sourced from~\citep{hu2024refchecker} and consists of tasks such as text summarization, closed-QA, and information extraction. Similar to RAGTruth, Dolly (AC) lacks hallucination annotations for responses generated by certain LLMs, particularly those without access to accurate context. To fill this gap, the same team of three qualified annotators was tasked with manually evaluating the RAG responses to determine if they contained hallucinations, focusing on the alignment between the generated responses and the external context.

In both cases, the manual annotation process involved cross-verifying the generated content with the provided external context to detect discrepancies or factual inconsistencies that would indicate hallucinations.

\section{Details about Baseline Models}
\label{sec:baseline_de}
This section provides details on the baseline models and hallucination detection methods used in our experiments. We categorize the methods into three groups based on their approach to leveraging parametric knowledge and external context: Parametric Confounded by External (PCE), External Confounded by Parametric (ECP), and Mixed Parametric and External (MPE).

\textbf{(1) Parametric Confounded by External (PCE):}
\begin{itemize}
    \item \textbf{EigenScore:} EigenScore measures the semantic consistency in the embedding space. Higher EigenScores suggest a higher likelihood of hallucinations, as they indicate greater semantic divergence~\citep{cheninside}.
    \item \textbf{SEP:} SEP (Semantic Entropy Probe) uses linear probes trained on the hidden states of LLMs to detect hallucinations by analyzing the semantic entropy of the tokens before generation~\citep{han2024semantic}.
    \item \textbf{SAPLMA:} SAPLMA trains a classifier on LLM activation values to detect hallucinations. It captures internal signals from the LLM's hidden layers to identify when the model might generate a hallucination~\citep{azaria2023internal}.
    \item \textbf{ITI:} Inference-Time Intervention (ITI) analyzes attention head activations and uses a binary classifier to predict hallucinations by studying the relationship between heads and task performance~\citep{li2024inference}.
\end{itemize}

\textbf{(2) External Confounded by Parametric (ECP):}
\begin{itemize}
    \item \textbf{Prompt:} This method uses the prompts provided in the RAGTruth~\citep{niu-etal-2024-ragtruth} to evaluate whether the LLM-generated responses are hallucinations by comparing them against the ground truth using GPT-4-o-mini~\citep{niu-etal-2024-ragtruth}.
    \item \textbf{LMvLM:} LMvLM employs a multi-turn interaction between two language models(GPT-4-o-mini vs. Backbone LLM) to discover inconsistencies by having them cross-examine each other’s responses~\citep{cohen2023lm}.
    \item \textbf{ChainPoll:} ChainPoll uses GPT-4-o-mini to determine if a completion contains hallucinations through a carefully designed prompt. The evaluation is repeated multiple times (typically five), and the final hallucination score is calculated as the ratio of "yes" answers to the total number of responses~\citep{friel2023chainpoll}.
    \item \textbf{RAGAS:} RAGAS checks the faithfulness of the generated response by breaking down sentences into shorter assertions and verifying each against the context, using GPT-4-o-mini to calculate a faithfulness score as the ratio of supported statements~\citep{es-etal-2024-ragas}.
    \item \textbf{Trulens:} Trulens assesses the overlap of information between the context and the generated response using GPT-4-o-mini, assigning a groundedness score between 0 and 10 based on the degree of overlap~\citep{trulens}.
    \item \textbf{RefCheck:} Similar to RAGAS, RefCheck extracts knowledge graphs from the generated responses and evaluates whether the knowledge graphs align with the external context~\citep{hu2024refchecker}.
    \item \textbf{P(True):} P(True) measures the uncertainty of the generated claim by querying the LLM itself on the truthfulness of its generated response. The confidence score is calculated as the probability of the first token being "True"~\citep{kadavath2022language}.
\end{itemize}

\textbf{(3) Mixed Parametric and External (MPE):}
\begin{itemize}
    \item \textbf{SelfCheckGPT:} SelfCheckGPT uses a zero-resource, sampling-based approach where multiple reference responses are checked by GPT-4-o-mini for consistency with the generated answer~\citep{manakul2023selfcheckgpt}.
    \item \textbf{LN-Entropy:} Length-Normalized Entropy measures sequence-level uncertainty across multiple generations, using entropy normalized by sequence length to detect hallucinations~\citep{malinin2020uncertainty}.
    \item \textbf{Energy:} Energy-based OOD detection identifies hallucinations by analyzing the uncertainty in the generated response using energy functions. Higher energy suggests a higher likelihood of hallucinations~\citep{liu2020energy}.
    \item \textbf{Focus:} Focus enhances uncertainty-based hallucination detection by focusing on key informative tokens, preceding words, and token properties, simulating human factuality checking~\citep{zhang2023enhancing}.
    \item \textbf{Perplexity:} This method uses the perplexity of the LLM-generated response to detect hallucinations. A higher perplexity indicates greater uncertainty and a higher likelihood of hallucinations~\citep{ren2022out}.
\end{itemize}

\section{Implementation Details.}
\label{sec:imp}

We run all the experiments on machines equipped with NVIDIA V100 GPUs and 52-core Intel(R) Xeon(R) Gold 6230R CPUs at 2.10GHz. We utilize the Huggingface Transformers package to conduct experiments. During the decoding of responses from the language models, we employ greedy search to generate responses. The remaining parameters follow the models' default settings. For RAGTruth, we use the validation set to select the hyperparameters. For Dolly (AC), we use two-fold validation to select the hyperparameters. For the baselines, we perform hyperparameter tuning within the range provided by the original works.
For ReDeEP(Chunk) on Dolly (AC), on Llama2-7B, we select the top-7 scoring Copying Head and top-3 FFN layers with $\alpha = 1$ and $\beta = 1.6$, as described in~\Secref{sec:empirical}. On Llama2-13B, we select the top-11 scoring Copying Head and top-3 FFN layers with $\alpha = 1$ and $\beta = 0.2$. On Llama3-8B, we select the top-1 scoring Copying Head and top-1 FFN layers with $\alpha = 1$ and $\beta = 0.1$, as described in~\Secref{sec:empirical}. For ReDeEP(Chunk) on RAGTruth, on Llama2-7B, we select the top-3 scoring Copying Head and top-4 FFN layers with $\alpha = 1$ and $\beta = 0.6$. On Llama2-13B, we select the top-9 scoring Copying Head and top-3 FFN layers with $\alpha = 1$ and $\beta = 1.8$. On Llama3-8B, we select the top-2 scoring Copying Head and top-5 FFN layers with $\alpha = 1$ and $\beta = 1.2$.
For ReDeEP(Token) on Dolly (AC), on Llama2-7B, we select the top-4 scoring Copying Head and top-3 FFN layers with $\alpha = 1$ and $\beta = 0.2$, as described in~\Secref{sec:empirical}. On Llama2-13B, we select the top-4 scoring Copying Head and top-5 FFN layers with $\alpha = 1$ and $\beta = 0.6$. On Llama3-8B, we select the top-1 scoring Copying Head and top-1 FFN layers with $\alpha = 1$ and $\beta = 0.1$, as described in~\Secref{sec:empirical}. For ReDeEP(Token) on RAGTruth, on Llama2-7B, we select the top-1 scoring Copying Head and top-10 FFN layers with $\alpha = 1$ and $\beta = 0.2$. On Llama2-13B, we select the top-2 scoring Copying Head and top-17 FFN layers with $\alpha = 1$ and $\beta = 0.6$. On Llama3-8B, we select the top-3 scoring Copying Head and top-30 FFN layers with $\alpha = 1$ and $\beta = 0.4$.
For \textbf{AARF} on RAGTruth, for Llama2-7B and Llama2-13B, we select $\alpha_1 = 5$, $\beta_1 = 0.2$, and $\tau = 0.4$, while for Llama3-8B, we use $\alpha_1 = 5$, $\beta_1 = 0.2$, and $\tau = 0.0$. On Dolly (AC), for Llama2-7B, Llama2-13B, and Llama3-8B, we select $\alpha_1 = 2$, $\beta_1 = 0.5$, and $\tau = 0.6$.
As proposed in \Secref{sec:chunk_det}, ReDeEP (Chunk) requires segmenting the retrieved documents and responses from the benchmark. For this, we utilized LangChain~\footnote{\url{https://www.langchain.com/}}, a popular open-source toolkit, and applied the RecursiveCharacterTextSplitter for the segmentation process.
We use BGE embeddings~\cite{xiao2023c} for ReDeEP(chunk), which is the sota embedding model.
The Llama2-7B can be downloaded from~\url{https://huggingface.co/meta-llama/Llama-2-7b-chat-hf}.
The Llama2-13B can be downloaded from~\url{https://huggingface.co/meta-llama/Llama-2-13b-chat-hf}.
The Llama3-8B can be downloaded from~\url{https://huggingface.co/meta-llama/Meta-Llama-3-8B-Instruct}.
The BGE embeddings can be downloaded from~\url{https://huggingface.co/BAAI/bge-base-en-v1.5}.

\section{Ablation Study}
\label{sec:ablation}
\begin{table}[t]
    \centering
    \resizebox{0.99\linewidth}{!}{
        \renewcommand\arraystretch{1.1}
    \centering
    \setlength{\tabcolsep}{1.5mm}
    \begin{tabular}{lc|cc|lc|cc}
    \toprule
         \multicolumn{8}{c}{\textbf{RAGTruth}}\\ \midrule
         \multicolumn{2}{c}{\textbf{ReDeEP (Token)}}  & AUC & PCC & \multicolumn{2}{c}{\textbf{ReDeEP (Chunk)}} & AUC & PCC  \\ \midrule
        \multirow{3}{*}{\textbf{LLaMA2-7B}} & Only PKS & 0.6950 & 0.3327 & \multirow{3}{*}{\textbf{LLaMA2-7B}}  & Only PKS & 0.6180 & 0.2103 \\ 
        &Only ECS & 0.7234 & 0.3779 & &Only ECS & 0.7098 & 0.3944 \\ 
        &Full & 0.7325 & 0.3979 & &Full & 0.7458 & 0.4203 \\ \midrule
        \multirow{3}{*}{\textbf{LLaMA2-13B}}  & Only PKS & 0.7214 & 0.3682 & \multirow{3}{*}{\textbf{LLaMA2-13B}} & Only PKS & 0.6614 & 0.2566 \\ 
        &Only ECS & 0.8040 & 0.5201 & &Only ECS & 0.7231 & 0.3922 \\ 
        &Full & 0.8181 & 0.5478 & &Full & 0.8244 & 0.5566 \\ \midrule
        \multirow{3}{*}{\textbf{LLaMA3-8B}} & Only PKS & 0.6102 & 0.1085 & \multirow{3}{*}{\textbf{LLaMA3-8B}} & Only PKS & 0.6082 & 0.1695 \\ 
        &Only ECS & 0.7336 & 0.4312 & &Only ECS & 0.6923 & 0.3102\\ 
        &Full & 0.7522 & 0.4493 & &Full & 0.7285 & 0.3964 \\ \midrule \midrule
        \multicolumn{8}{c}{\textbf{Dolly (AC)}} \\ \midrule
        \multicolumn{2}{c}{\textbf{ReDeEP (Token)}}  & AUC & PCC & \multicolumn{2}{c}{\textbf{ReDeEP (Chunk)}}& AUC & PCC  \\ \midrule
        \multirow{3}{*}{\textbf{LLaMA2-7B}} & Only PKS & 0.6671 & 0.2374 & \multirow{3}{*}{\textbf{LLaMA2-7B}}  & Only PKS & 0.6383 & 0.2115 \\ 
        & Only ECS & 0.6629 & 0.2852 & & Only ECS & 0.7552 & 0.4478  \\ 
        & Full & 0.6884 & 0.3266 & & Full & 0.7949 & 0.5136 \\ \midrule
        \multirow{3}{*}{\textbf{LLaMA2-13B}} & Only PKS & 0.6639 & 0.2891 & \multirow{3}{*}{\textbf{LLaMA2-13B}} & Only PKS & 0.6790 & 0.2883 \\ 
        & Only ECS & 0.6856 & 0.3107 & & Only ECS & 0.7383 & 0.4338 \\ 
        & Full & 0.7226 & 0.3776 & & Full & 0.8420 & 0.5902\\ \midrule
        \multirow{3}{*}{\textbf{LLaMA3-8B}} & Only PKS & 0.6329 & 0.2300 & \multirow{3}{*}{\textbf{LLaMA3-8B}} & Only PKS & 0.7334 & 0.3503 \\ 
        & Only ECS & 0.6105 & 0.1556 & & Only ECS & 0.6166 & 0.2624 \\ 
        & Full & 0.6701 & 0.2421 & & Full & 0.7354 & 0.3652\\ \bottomrule
    \end{tabular}}
    \caption{Ablation Study of ReDeEP.}
    \label{tab:ablation}
\end{table}

As shown in Table~\ref{tab:ablation}, when performing RAG hallucination detection, using only the Parametric Knowledge Score (Only PKS) or only the External Context Score (Only ECS) does not achieve the same performance as the Full ReDeEP model. This validates the effectiveness of employing multivariate regression, where both PKS and ECS are used simultaneously as covariates. According to the analysis in ~\Secref{sec:intro}, using Only PKS or Only ECS introduces confounding issues, leading to a decrease in performance. This explains why both Only ECS and Only PKS yield lower results compared to Full ReDeEP.

\section{Efficiency Analysis}
\label{sec:eff}
\begin{figure}[ht]
    \centering
    \includegraphics[width=\textwidth]{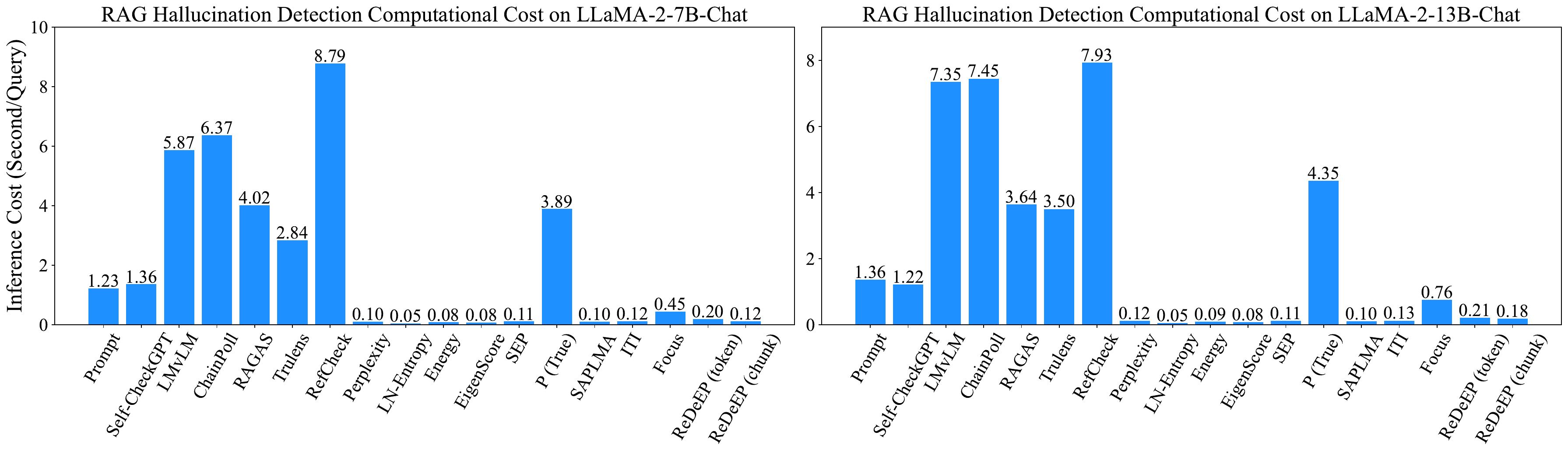}
    \caption{Comparison of the detection efficiency of different hallucination detection methods on LLaMA-2-7B-Chat and LLaMA-2-13B-Chat.}
    \label{fig:efficacy_analysis}
\end{figure}

As shown in Figure~\ref{fig:efficacy_analysis}, ReDeEP (chunk) is more efficient than ReDeEP (token), confirming the superior time efficiency of our proposed chunk-level hallucination detection method. Additionally, methods that do not rely on external models (e.g., ITI, ReDeEP, etc.) exhibit significantly higher efficiency compared to those using external models (e.g., RefCheck, Chainpoll, etc.). 
ReDeEP is positioned at a high-efficiency level among hallucination detection models, validating its feasibility for industrial applications. Considering that the main time consumption of AARF lies in ReDeEP (token), the relatively low time consumption of ReDeEP (token) further validates the efficiency of AARF.

\section{Prompt for truthfulness evaluation}
\label{sec:prompt_truthful_eval}

In the Truthful RAG Generation Experiments, we used the following prompt to assess the truthfulness of RAG generation from AARF and baseline models:

\begin{tcolorbox}[colback=gray!20, colframe=gray, coltitle=white, fonttitle=\bfseries]

You are tasked with evaluating two answers generated by different models: AARF and Baseline. Your goal is to determine which answer is more truthful based on the provided content and query. Please carefully compare the truthfulness of both answers. Follow the specific response format provided below.

\textbf{Query:} \{\textit{String}\}

\textbf{External Context:} \{\textit{String}\}

\textbf{AARF answer:} \{\textit{String}\}

\textbf{Baseline answer:} \{\textit{String}\}

Respond in the following format: 

\{
\{

"win\_model": "AARF (AARF's answer is more truthful) or Baseline (Baseline's answer is more truthful) or Tie (Both answers are equally truthful. Generally not needed. Choose this only if no distinction can be made)", 

"reason": "Provide the reason for selecting the winning model's answer as more truthful."

\}\}
\end{tcolorbox}

\end{document}

%% file: math_commands.tex
\usepackage{amsmath,amsfonts,bm}

\def\Figref#1{Figure~\ref{#1}}

\def\Secref#1{Section~\ref{#1}}

\def\eqref#1{equation~\ref{#1}}
\def\Eqref#1{Equation~\ref{#1}}

\def\1{\bm{1}}

\def\ermE{{\textnormal{E}}}

\def\ermH{{\textnormal{H}}}

\def\ermP{{\textnormal{P}}}

\DeclareMathAlphabet{\mathsfit}{\encodingdefault}{\sfdefault}{m}{sl}
\SetMathAlphabet{\mathsfit}{bold}{\encodingdefault}{\sfdefault}{bx}{n}













%% file: iclr2025_conference.bbl
\begin{thebibliography}{59}
\providecommand{\natexlab}[1]{#1}
\providecommand{\url}[1]{\texttt{#1}}
\expandafter\ifx\csname urlstyle\endcsname\relax
  \providecommand{\doi}[1]{doi: #1}\else
  \providecommand{\doi}{doi: \begingroup \urlstyle{rm}\Url}\fi

\bibitem[Achiam et~al.(2023)Achiam, Adler, Agarwal, Ahmad, Akkaya, Aleman, Almeida, Altenschmidt, Altman, Anadkat, et~al.]{achiam2023gpt}
Josh Achiam, Steven Adler, Sandhini Agarwal, Lama Ahmad, Ilge Akkaya, Florencia~Leoni Aleman, Diogo Almeida, Janko Altenschmidt, Sam Altman, Shyamal Anadkat, et~al.
\newblock Gpt-4 technical report.
\newblock \emph{arXiv preprint arXiv:2303.08774}, 2023.

\bibitem[Azaria \& Mitchell(2023)Azaria and Mitchell]{azaria2023internal}
Amos Azaria and Tom Mitchell.
\newblock The internal state of an llm knows when it’s lying.
\newblock In \emph{Findings of the Association for Computational Linguistics: EMNLP 2023}, pp.\  967--976, 2023.

\bibitem[Bell(1965)]{bell1965gershgorin}
Howard~E Bell.
\newblock Gershgorin's theorem and the zeros of polynomials.
\newblock \emph{The American Mathematical Monthly}, 72\penalty0 (3):\penalty0 292--295, 1965.

\bibitem[Bengio et~al.(2013)Bengio, Courville, and Vincent]{bengio2013representation}
Yoshua Bengio, Aaron Courville, and Pascal Vincent.
\newblock Representation learning: A review and new perspectives.
\newblock \emph{IEEE transactions on pattern analysis and machine intelligence}, 35\penalty0 (8):\penalty0 1798--1828, 2013.

\bibitem[Chen et~al.(2024{\natexlab{a}})Chen, Liu, Chen, Gu, Wu, Tao, Fu, and Ye]{chen2024inside}
Chao Chen, Kai Liu, Ze~Chen, Yi~Gu, Yue Wu, Mingyuan Tao, Zhihang Fu, and Jieping Ye.
\newblock {INSIDE}: {LLM}s' internal states retain the power of hallucination detection.
\newblock In \emph{The Twelfth International Conference on Learning Representations}, 2024{\natexlab{a}}.
\newblock URL \url{https://openreview.net/forum?id=Zj12nzlQbz}.

\bibitem[Chen et~al.(2024{\natexlab{b}})Chen, Liu, Chen, Gu, Wu, Tao, Fu, and Ye]{cheninside}
Chao Chen, Kai Liu, Ze~Chen, Yi~Gu, Yue Wu, Mingyuan Tao, Zhihang Fu, and Jieping Ye.
\newblock Inside: Llms' internal states retain the power of hallucination detection.
\newblock In \emph{The Twelfth International Conference on Learning Representations}, 2024{\natexlab{b}}.

\bibitem[Chuang et~al.(2024)Chuang, Xie, Luo, Kim, Glass, and He]{chuang2024dola}
Yung-Sung Chuang, Yujia Xie, Hongyin Luo, Yoon Kim, James~R. Glass, and Pengcheng He.
\newblock Dola: Decoding by contrasting layers improves factuality in large language models.
\newblock In \emph{The Twelfth International Conference on Learning Representations}, 2024.
\newblock URL \url{https://openreview.net/forum?id=Th6NyL07na}.

\bibitem[Chyzhyk et~al.(2022)Chyzhyk, Varoquaux, Milham, and Thirion]{chyzhyk2022remove}
Darya Chyzhyk, Ga{\"e}l Varoquaux, Michael Milham, and Bertrand Thirion.
\newblock How to remove or control confounds in predictive models, with applications to brain biomarkers.
\newblock \emph{GigaScience}, 11:\penalty0 giac014, 2022.

\bibitem[Clark et~al.(2019)Clark, Khandelwal, Levy, and Manning]{clark-etal-2019-bert}
Kevin Clark, Urvashi Khandelwal, Omer Levy, and Christopher~D. Manning.
\newblock What does {BERT} look at? an analysis of {BERT}{'}s attention.
\newblock In Tal Linzen, Grzegorz Chrupa{\l}a, Yonatan Belinkov, and Dieuwke Hupkes (eds.), \emph{Proceedings of the 2019 ACL Workshop BlackboxNLP: Analyzing and Interpreting Neural Networks for NLP}, pp.\  276--286, Florence, Italy, August 2019. Association for Computational Linguistics.
\newblock \doi{10.18653/v1/W19-4828}.
\newblock URL \url{https://aclanthology.org/W19-4828}.

\bibitem[Cohen et~al.(2023)Cohen, Hamri, Geva, and Globerson]{cohen2023lm}
Roi Cohen, May Hamri, Mor Geva, and Amir Globerson.
\newblock Lm vs lm: Detecting factual errors via cross examination.
\newblock \emph{arXiv preprint arXiv:2305.13281}, 2023.

\bibitem[Dai et~al.(2022)Dai, Dong, Hao, Sui, Chang, and Wei]{dai2022knowledge}
Damai Dai, Li~Dong, Yaru Hao, Zhifang Sui, Baobao Chang, and Furu Wei.
\newblock Knowledge neurons in pretrained transformers.
\newblock In \emph{Proceedings of the 60th Annual Meeting of the Association for Computational Linguistics (Volume 1: Long Papers)}, pp.\  8493--8502, 2022.

\bibitem[Dubey et~al.(2024)Dubey, Jauhri, Pandey, Kadian, Al-Dahle, Letman, Mathur, Schelten, Yang, Fan, et~al.]{dubey2024llama}
Abhimanyu Dubey, Abhinav Jauhri, Abhinav Pandey, Abhishek Kadian, Ahmad Al-Dahle, Aiesha Letman, Akhil Mathur, Alan Schelten, Amy Yang, Angela Fan, et~al.
\newblock The llama 3 herd of models.
\newblock \emph{arXiv preprint arXiv:2407.21783}, 2024.

\bibitem[Elhage et~al.(2021)Elhage, Nanda, Olsson, Henighan, Joseph, Mann, Askell, Bai, Chen, Conerly, DasSarma, Drain, Ganguli, Hatfield-Dodds, Hernandez, Jones, Kernion, Lovitt, Ndousse, Amodei, Brown, Clark, Kaplan, McCandlish, and Olah]{transformercircuits2021}
Nelson Elhage, Neel Nanda, Catherine Olsson, Tom Henighan, Nicholas Joseph, Ben Mann, Amanda Askell, Yuntao Bai, Anna Chen, Tom Conerly, Nova DasSarma, Dawn Drain, Deep Ganguli, Zac Hatfield-Dodds, Danny Hernandez, Andy Jones, Jackson Kernion, Liane Lovitt, Kamal Ndousse, Dario Amodei, Tom Brown, Jack Clark, Jared Kaplan, Sam McCandlish, and Chris Olah.
\newblock A mathematical framework for transformer circuits.
\newblock Transformer Circuits Thread, 2021.
\newblock URL \url{https://transformer-circuits.pub/2021/framework/index.html}.

\bibitem[Es et~al.(2024)Es, James, Espinosa~Anke, and Schockaert]{es-etal-2024-ragas}
Shahul Es, Jithin James, Luis Espinosa~Anke, and Steven Schockaert.
\newblock {RAGA}s: Automated evaluation of retrieval augmented generation.
\newblock In Nikolaos Aletras and Orphee De~Clercq (eds.), \emph{Proceedings of the 18th Conference of the European Chapter of the Association for Computational Linguistics: System Demonstrations}, pp.\  150--158, St. Julians, Malta, March 2024. Association for Computational Linguistics.
\newblock URL \url{https://aclanthology.org/2024.eacl-demo.16}.

\bibitem[Fan et~al.(2024)Fan, Ding, Ning, Wang, Li, Yin, Chua, and Li]{fan2024survey}
Wenqi Fan, Yujuan Ding, Liangbo Ning, Shijie Wang, Hengyun Li, Dawei Yin, Tat-Seng Chua, and Qing Li.
\newblock A survey on rag meeting llms: Towards retrieval-augmented large language models.
\newblock In \emph{Proceedings of the 30th ACM SIGKDD Conference on Knowledge Discovery and Data Mining}, pp.\  6491--6501, 2024.

\bibitem[Ferrando \& Voita(2024)Ferrando and Voita]{ferrando2024information}
Javier Ferrando and Elena Voita.
\newblock Information flow routes: Automatically interpreting language models at scale.
\newblock \emph{arXiv preprint arXiv:2403.00824}, 2024.

\bibitem[Ferrando et~al.(2024)Ferrando, Sarti, Bisazza, and Costa-juss{\`a}]{ferrando2024primer}
Javier Ferrando, Gabriele Sarti, Arianna Bisazza, and Marta~R Costa-juss{\`a}.
\newblock A primer on the inner workings of transformer-based language models.
\newblock \emph{arXiv preprint arXiv:2405.00208}, 2024.

\bibitem[Finardi et~al.(2024)Finardi, Avila, Castaldoni, Gengo, Larcher, Piau, Costa, and Carid{\'a}]{finardi2024chronicles}
Paulo Finardi, Leonardo Avila, Rodrigo Castaldoni, Pedro Gengo, Celio Larcher, Marcos Piau, Pablo Costa, and Vinicius Carid{\'a}.
\newblock The chronicles of rag: The retriever, the chunk and the generator.
\newblock \emph{arXiv preprint arXiv:2401.07883}, 2024.

\bibitem[Friel \& Sanyal(2023)Friel and Sanyal]{friel2023chainpoll}
Robert Friel and Atindriyo Sanyal.
\newblock Chainpoll: A high efficacy method for llm hallucination detection.
\newblock \emph{arXiv preprint arXiv:2310.18344}, 2023.

\bibitem[Gao et~al.(2023)Gao, Xiong, Gao, Jia, Pan, Bi, Dai, Sun, and Wang]{gao2023retrieval}
Yunfan Gao, Yun Xiong, Xinyu Gao, Kangxiang Jia, Jinliu Pan, Yuxi Bi, Yi~Dai, Jiawei Sun, and Haofen Wang.
\newblock Retrieval-augmented generation for large language models: A survey.
\newblock \emph{arXiv preprint arXiv:2312.10997}, 2023.

\bibitem[Geva et~al.(2021)Geva, Schuster, Berant, and Levy]{geva2021transformer}
Mor Geva, Roei Schuster, Jonathan Berant, and Omer Levy.
\newblock Transformer feed-forward layers are key-value memories.
\newblock In \emph{Proceedings of the 2021 Conference on Empirical Methods in Natural Language Processing}, pp.\  5484--5495, 2021.

\bibitem[Han et~al.(2024)Han, Kossen, Razzak, Schut, Malik, and Gal]{han2024semantic}
Jiatong Han, Jannik Kossen, Muhammed Razzak, Lisa Schut, Shreshth~A Malik, and Yarin Gal.
\newblock Semantic entropy probes: Robust and cheap hallucination detection in llms.
\newblock In \emph{ICML 2024 Workshop on Foundation Models in the Wild}, 2024.

\bibitem[Hanna et~al.(2024)Hanna, Liu, and Variengien]{hanna2024does}
Michael Hanna, Ollie Liu, and Alexandre Variengien.
\newblock How does gpt-2 compute greater-than?: Interpreting mathematical abilities in a pre-trained language model.
\newblock \emph{Advances in Neural Information Processing Systems}, 36, 2024.

\bibitem[Hu et~al.(2024)Hu, Ru, Qiu, Guo, Zhang, Xu, Luo, Liu, Zhang, and Zhang]{hu2024refchecker}
Xiangkun Hu, Dongyu Ru, Lin Qiu, Qipeng Guo, Tianhang Zhang, Yang Xu, Yun Luo, Pengfei Liu, Yue Zhang, and Zheng Zhang.
\newblock Refchecker: Reference-based fine-grained hallucination checker and benchmark for large language models.
\newblock \emph{arXiv preprint arXiv:2405.14486}, 2024.

\bibitem[Huang et~al.(2023)Huang, Yu, Ma, Zhong, Feng, Wang, Chen, Peng, Feng, Qin, et~al.]{huang2023survey}
Lei Huang, Weijiang Yu, Weitao Ma, Weihong Zhong, Zhangyin Feng, Haotian Wang, Qianglong Chen, Weihua Peng, Xiaocheng Feng, Bing Qin, et~al.
\newblock A survey on hallucination in large language models: Principles, taxonomy, challenges, and open questions.
\newblock \emph{arXiv preprint arXiv:2311.05232}, 2023.

\bibitem[Kadavath et~al.(2022)Kadavath, Conerly, Askell, Henighan, Drain, Perez, Schiefer, Hatfield-Dodds, DasSarma, Tran-Johnson, et~al.]{kadavath2022language}
Saurav Kadavath, Tom Conerly, Amanda Askell, Tom Henighan, Dawn Drain, Ethan Perez, Nicholas Schiefer, Zac Hatfield-Dodds, Nova DasSarma, Eli Tran-Johnson, et~al.
\newblock Language models (mostly) know what they know.
\newblock \emph{arXiv preprint arXiv:2207.05221}, 2022.

\bibitem[Kahlert et~al.(2017)Kahlert, Gribsholt, Gammelager, Dekkers, and Luta]{kahlert2017control}
Johnny Kahlert, Sigrid~Bjerge Gribsholt, Henrik Gammelager, Olaf~M Dekkers, and George Luta.
\newblock Control of confounding in the analysis phase--an overview for clinicians.
\newblock \emph{Clinical epidemiology}, pp.\  195--204, 2017.

\bibitem[Kobayashi et~al.(2021)Kobayashi, Kuribayashi, Yokoi, and Inui]{kobayashi2021incorporating}
Goro Kobayashi, Tatsuki Kuribayashi, Sho Yokoi, and Kentaro Inui.
\newblock Incorporating residual and normalization layers into analysis of masked language models.
\newblock In \emph{Proceedings of the 2021 Conference on Empirical Methods in Natural Language Processing}, pp.\  4547--4568, 2021.

\bibitem[Li et~al.(2024)Li, Patel, Vi{\'e}gas, Pfister, and Wattenberg]{li2024inference}
Kenneth Li, Oam Patel, Fernanda Vi{\'e}gas, Hanspeter Pfister, and Martin Wattenberg.
\newblock Inference-time intervention: Eliciting truthful answers from a language model.
\newblock \emph{Advances in Neural Information Processing Systems}, 36, 2024.

\bibitem[Liu et~al.(2020)Liu, Wang, Owens, and Li]{liu2020energy}
Weitang Liu, Xiaoyun Wang, John Owens, and Yixuan Li.
\newblock Energy-based out-of-distribution detection.
\newblock \emph{Advances in neural information processing systems}, 33:\penalty0 21464--21475, 2020.

\bibitem[Luo et~al.(2024)Luo, Qin, Liu, Xiao, Zhao, and Liu]{luo2024foundations}
Kun Luo, Minghao Qin, Zheng Liu, Shitao Xiao, Jun Zhao, and Kang Liu.
\newblock Large language models as foundations for next-gen dense retrieval: A comprehensive empirical assessment, 2024.
\newblock URL \url{https://arxiv.org/abs/2408.12194}.

\bibitem[Magesh et~al.(2024)Magesh, Surani, Dahl, Suzgun, Manning, and Ho]{magesh2024hallucination}
Varun Magesh, Faiz Surani, Matthew Dahl, Mirac Suzgun, Christopher~D Manning, and Daniel~E Ho.
\newblock Hallucination-free? assessing the reliability of leading ai legal research tools.
\newblock \emph{arXiv preprint arXiv:2405.20362}, 2024.

\bibitem[Malinin \& Gales(2020)Malinin and Gales]{malinin2020uncertainty}
Andrey Malinin and Mark Gales.
\newblock Uncertainty estimation in autoregressive structured prediction.
\newblock \emph{arXiv preprint arXiv:2002.07650}, 2020.

\bibitem[Manakul et~al.(2023)Manakul, Liusie, and Gales]{manakul2023selfcheckgpt}
Potsawee Manakul, Adian Liusie, and Mark Gales.
\newblock Selfcheckgpt: Zero-resource black-box hallucination detection for generative large language models.
\newblock In \emph{Proceedings of the 2023 Conference on Empirical Methods in Natural Language Processing}, pp.\  9004--9017, 2023.

\bibitem[McDougall et~al.(2023)McDougall, Conmy, Rushing, McGrath, and Nanda]{mcdougall2023copy}
Callum McDougall, Arthur Conmy, Cody Rushing, Thomas McGrath, and Neel Nanda.
\newblock Copy suppression: Comprehensively understanding an attention head.
\newblock \emph{arXiv preprint arXiv:2310.04625}, 2023.

\bibitem[Meng et~al.(2022)Meng, Bau, Andonian, and Belinkov]{meng2022locating}
Kevin Meng, David Bau, Alex Andonian, and Yonatan Belinkov.
\newblock Locating and editing factual associations in gpt.
\newblock \emph{Advances in Neural Information Processing Systems}, 35:\penalty0 17359--17372, 2022.

\bibitem[Neuberg(2003)]{neuberg2003causality}
Leland~Gerson Neuberg.
\newblock Causality: models, reasoning, and inference, by judea pearl, cambridge university press, 2000.
\newblock \emph{Econometric Theory}, 19\penalty0 (4):\penalty0 675--685, 2003.

\bibitem[Niu et~al.(2024)Niu, Wu, Zhu, Xu, Shum, Zhong, Song, and Zhang]{niu-etal-2024-ragtruth}
Cheng Niu, Yuanhao Wu, Juno Zhu, Siliang Xu, KaShun Shum, Randy Zhong, Juntong Song, and Tong Zhang.
\newblock {RAGT}ruth: A hallucination corpus for developing trustworthy retrieval-augmented language models.
\newblock In \emph{Proceedings of the 62nd Annual Meeting of the Association for Computational Linguistics (Volume 1: Long Papers)}, Bangkok, Thailand, August 2024. Association for Computational Linguistics.

\bibitem[nostalgebraist(2020)]{nostalgebraist2020}
nostalgebraist.
\newblock Interpreting {GPT}: the logit lens.
\newblock \emph{AI Alignment Forum}, 2020.
\newblock URL \url{https://www.alignmentforum.org/posts/AcKRB8wDpdaN6v6ru/interpreting-gpt-the-logit-lens}.

\bibitem[Olsson et~al.(2022)Olsson, Elhage, Nanda, Joseph, DasSarma, Henighan, Mann, Askell, Bai, Chen, et~al.]{olsson2022context}
Catherine Olsson, Nelson Elhage, Neel Nanda, Nicholas Joseph, Nova DasSarma, Tom Henighan, Ben Mann, Amanda Askell, Yuntao Bai, Anna Chen, et~al.
\newblock In-context learning and induction heads.
\newblock \emph{arXiv preprint arXiv:2209.11895}, 2022.

\bibitem[Pearl(2009)]{pearl2009causality}
J~Pearl.
\newblock \emph{Causality}.
\newblock Cambridge university press, 2009.

\bibitem[{PyTorch}(2023)]{pytorch_nllloss}
{PyTorch}.
\newblock torch.nn.nllloss.
\newblock \url{https://pytorch.org/docs/stable/generated/torch.nn.NLLLoss.html}, 2023.
\newblock Accessed: 2023-09-20.

\bibitem[Ren et~al.(2022)Ren, Luo, Zhao, Krishna, Saleh, Lakshminarayanan, and Liu]{ren2022out}
Jie Ren, Jiaming Luo, Yao Zhao, Kundan Krishna, Mohammad Saleh, Balaji Lakshminarayanan, and Peter~J Liu.
\newblock Out-of-distribution detection and selective generation for conditional language models.
\newblock In \emph{The Eleventh International Conference on Learning Representations}, 2022.

\bibitem[Schuster et~al.(2022)Schuster, Fisch, Gupta, Dehghani, Bahri, Tran, Tay, and Metzler]{schuster2022confident}
Tal Schuster, Adam Fisch, Jai Gupta, Mostafa Dehghani, Dara Bahri, Vinh~Q. Tran, Yi~Tay, and Donald Metzler.
\newblock Confident adaptive language modeling.
\newblock In Alice~H. Oh, Alekh Agarwal, Danielle Belgrave, and Kyunghyun Cho (eds.), \emph{Advances in Neural Information Processing Systems}, 2022.
\newblock URL \url{https://openreview.net/forum?id=uLYc4L3C81A}.

\bibitem[Shuster et~al.(2021)Shuster, Poff, Chen, Kiela, and Weston]{shuster2021retrieval}
Kurt Shuster, Spencer Poff, Moya Chen, Douwe Kiela, and Jason Weston.
\newblock Retrieval augmentation reduces hallucination in conversation.
\newblock In \emph{Findings of the Association for Computational Linguistics: EMNLP 2021}, pp.\  3784--3803, 2021.

\bibitem[Stuart(2010)]{stuart2010matching}
Elizabeth~A Stuart.
\newblock Matching methods for causal inference: A review and a look forward.
\newblock \emph{Statistical science: a review journal of the Institute of Mathematical Statistics}, 25\penalty0 (1):\penalty0 1, 2010.

\bibitem[Touvron et~al.(2023)Touvron, Martin, Stone, Albert, Almahairi, Babaei, Bashlykov, Batra, Bhargava, Bhosale, et~al.]{touvron2023llama}
Hugo Touvron, Louis Martin, Kevin Stone, Peter Albert, Amjad Almahairi, Yasmine Babaei, Nikolay Bashlykov, Soumya Batra, Prajjwal Bhargava, Shruti Bhosale, et~al.
\newblock Llama 2: Open foundation and fine-tuned chat models.
\newblock \emph{arXiv preprint arXiv:2307.09288}, 2023.

\bibitem[{Trulens}(2024)]{trulens}
{Trulens}.
\newblock Trulens: Evaluate and track llm applications, 2024.
\newblock URL \url{https://www.trulens.org/}.

\bibitem[Vaswani(2017)]{vaswani2017attention}
A~Vaswani.
\newblock Attention is all you need.
\newblock \emph{Advances in Neural Information Processing Systems}, 2017.

\bibitem[Vinutha et~al.(2018)Vinutha, Poornima, and Sagar]{vinutha2018detection}
HP~Vinutha, B~Poornima, and BM~Sagar.
\newblock Detection of outliers using interquartile range technique from intrusion dataset.
\newblock In \emph{Information and decision sciences: Proceedings of the 6th international conference on ficta}, pp.\  511--518. Springer, 2018.

\bibitem[Wadhwa et~al.(2024)Wadhwa, Seetharaman, Aggarwal, Ghosh, Basu, Srinivasan, Zhao, Chaudhari, and Aghazadeh]{wadhwa2024rags}
Hitesh Wadhwa, Rahul Seetharaman, Somyaa Aggarwal, Reshmi Ghosh, Samyadeep Basu, Soundararajan Srinivasan, Wenlong Zhao, Shreyas Chaudhari, and Ehsan Aghazadeh.
\newblock From rags to rich parameters: Probing how language models utilize external knowledge over parametric information for factual queries.
\newblock \emph{arXiv preprint arXiv:2406.12824}, 2024.

\bibitem[Wu et~al.(2024)Wu, Wang, Xiao, Peng, and Fu]{wu2024retrieval}
Wenhao Wu, Yizhong Wang, Guangxuan Xiao, Hao Peng, and Yao Fu.
\newblock Retrieval head mechanistically explains long-context factuality.
\newblock \emph{arXiv preprint arXiv:2404.15574}, 2024.

\bibitem[Xiao et~al.(2023)Xiao, Liu, Zhang, and Muennighof]{xiao2023c}
Shitao Xiao, Zheng Liu, Peitian Zhang, and Niklas Muennighof.
\newblock C-pack: Packaged resources to advance general chinese embedding.
\newblock \emph{arXiv preprint arXiv:2309.07597}, 2023.

\bibitem[Xu et~al.(2024)Xu, Qi, Wang, Wang, Zhang, and Xu]{xu2024knowledge}
Rongwu Xu, Zehan Qi, Cunxiang Wang, Hongru Wang, Yue Zhang, and Wei Xu.
\newblock Knowledge conflicts for llms: A survey.
\newblock \emph{arXiv preprint arXiv:2403.08319}, 2024.

\bibitem[Yu et~al.(2023)Yu, Merullo, and Pavlick]{yu2023characterizing}
Qinan Yu, Jack Merullo, and Ellie Pavlick.
\newblock Characterizing mechanisms for factual recall in language models.
\newblock In \emph{The 2023 Conference on Empirical Methods in Natural Language Processing}, 2023.

\bibitem[Zhang et~al.(2023)Zhang, Qiu, Guo, Deng, Zhang, Zhang, Zhou, Wang, and Fu]{zhang2023enhancing}
Tianhang Zhang, Lin Qiu, Qipeng Guo, Cheng Deng, Yue Zhang, Zheng Zhang, Chenghu Zhou, Xinbing Wang, and Luoyi Fu.
\newblock Enhancing uncertainty-based hallucination detection with stronger focus.
\newblock In \emph{Proceedings of the 2023 Conference on Empirical Methods in Natural Language Processing}, pp.\  915--932, 2023.

\bibitem[Zhang et~al.(2024)Zhang, Sheng, Zhou, Chen, Zheng, Cai, Song, Tian, R{\'e}, Barrett, et~al.]{zhang2024h2o}
Zhenyu Zhang, Ying Sheng, Tianyi Zhou, Tianlong Chen, Lianmin Zheng, Ruisi Cai, Zhao Song, Yuandong Tian, Christopher R{\'e}, Clark Barrett, et~al.
\newblock H2o: Heavy-hitter oracle for efficient generative inference of large language models.
\newblock \emph{Advances in Neural Information Processing Systems}, 36, 2024.

\bibitem[Zhou et~al.(2024)Zhou, Feng, Zhu, Qian, and Mao]{zhou2024unibias}
Hanzhang Zhou, Zijian Feng, Zixiao Zhu, Junlang Qian, and Kezhi Mao.
\newblock Unibias: Unveiling and mitigating llm bias through internal attention and ffn manipulation.
\newblock \emph{arXiv preprint arXiv:2405.20612}, 2024.

\bibitem[Zhu et~al.(2024)Zhu, Duan, Chen, Liu, Li, Feng, Lv, Cao, Chuanfu, Zhang, et~al.]{zhu2024near}
Qianchao Zhu, Jiangfei Duan, Chang Chen, Siran Liu, Xiuhong Li, Guanyu Feng, Xin Lv, Huanqi Cao, Xiao Chuanfu, Xingcheng Zhang, et~al.
\newblock Near-lossless acceleration of long context llm inference with adaptive structured sparse attention.
\newblock \emph{arXiv preprint arXiv:2406.15486}, 2024.

\end{thebibliography}
